\documentclass{article}

\usepackage[preprint,nonatbib]{neurips_2021}

\usepackage[utf8]{inputenc} 
\usepackage[T1]{fontenc}    
\usepackage{hyperref}       
\usepackage{url}            
\usepackage{booktabs}       
\usepackage{amsfonts}       
\usepackage{nicefrac}       
\usepackage{microtype}      
\usepackage{xcolor}         
\usepackage{amsmath}
\usepackage{algorithm}
\usepackage{algpseudocode}
\usepackage{bbm}
\usepackage{amsthm}
\newtheorem{theorem}{Theorem}
\usepackage{tabularx}
\usepackage{graphicx}
\usepackage{comment}

\makeatletter
\newenvironment{breakablealgorithm}
{
	\begin{center}
		\refstepcounter{algorithm}
		\hrule height.8pt depth0pt \kern2pt
		\renewcommand{\caption}[2][\relax]{
			{\raggedright\textbf{\fname@algorithm~\thealgorithm} ##2\par}%
			\ifx\relax##1\relax 
			\addcontentsline{loa}{algorithm}{\protect\numberline{\thealgorithm}##2}%
			\else 
			\addcontentsline{loa}{algorithm}{\protect\numberline{\thealgorithm}##1}%
			\fi
			\kern2pt\hrule\kern2pt
		}
	}{
		\kern2pt\hrule\relax
	\end{center}
}
\makeatother

\title{Fast TreeSHAP: Accelerating SHAP Value Computation for Trees}

\author{%
  Jilei Yang \\
  LinkedIn Corporation \\
  \texttt{jlyang@linkedin.com} \\
}

\begin{document}

\maketitle

\begin{abstract}
SHAP (SHapley Additive exPlanation) values are one of the leading tools for interpreting machine learning models, with strong theoretical guarantees (consistency, local accuracy) and a wide availability of implementations and use cases. Even though computing SHAP values takes exponential time in general, TreeSHAP takes polynomial time on tree-based models. While the speedup is significant, TreeSHAP can still dominate the computation time of industry-level machine learning solutions on datasets with millions or more entries, causing delays in post-hoc model diagnosis and interpretation service. In this paper we present two new algorithms, Fast TreeSHAP v1 and v2, designed to improve the computational efficiency of TreeSHAP for large datasets. We empirically find that Fast TreeSHAP v1 is 1.5x faster than TreeSHAP while keeping the memory cost unchanged. Similarly, Fast TreeSHAP v2 is 2.5x faster than TreeSHAP, at the cost of a slightly higher memory usage, thanks to the pre-computation of expensive TreeSHAP steps. We also show that Fast TreeSHAP v2 is well-suited for multi-time model interpretations, resulting in as high as 3x faster explanation of newly incoming samples.
\end{abstract}

\section{Introduction}
Predictive machine learning models are almost everywhere in industry, and complex models such as random forest, gradient boosted trees, and deep neural networks are being widely used due to their high prediction accuracy. However, interpreting predictions from these complex models remains an important challenge, and many times the interpretations at individual sample level are of the most interest \cite{ribeiro2016should}. There exist several state-of-the-art sample-level model interpretation approaches, e.g., SHAP \cite{lundberg2017unified}, LIME \cite{ribeiro2016should}, and Integrated Gradient \cite{sundararajan2017axiomatic}. Among them, SHAP (SHapley Additive exPlanation) calculates SHAP values quantifying the contribution of each feature to the model prediction, by incorporating concepts from game theory and local explanations. In contrast to other approaches, SHAP has been justified as the only consistent feature attribution approach with several unique properties which agree with human intuition. Due to its solid theoretical guarantees, SHAP becomes one of the top recommendations of model interpretation approaches in industry \cite{lundberg2018explainable,zhang2020clinically,agius2020machine,ariza2020explainability}.

There exist several variants of SHAP. The general version is KernelSHAP \cite{lundberg2017unified}, which is model-agnostic, and generally takes exponential time to compute the exact SHAP values. Its variants include TreeSHAP \cite{lundberg2018consistent,lundberg2019explainable} and DeepSHAP \cite{lundberg2017unified}, which are designed specifically for tree-based models (e.g., decision tree, random forest, gradient boosted trees) and neural-network models. In these variants, the special model structures may lead to potential improvements in computational efficiency. For example, the tree structure enables a polynomial time complexity for TreeSHAP. In this paper, we choose TreeSHAP for further exploration, as tree-based models are widespread in industry.

After looking into many TreeSHAP use cases, we find out that despite its algorithmic complexity improvement, computing SHAP values for large sample size (e.g., tens of millions of samples) or large model size (e.g., maximum tree depth $\geq10$) still remains a computational concern in practice. For example, explaining 20 million samples for a random forest model with 400 trees and maximum tree depth 12 can take as long as 15 hours even on a 100-core server (more details in Appendix \ref{subsec:real_life_scenario}). In fact, explaining (at least) tens of millions of samples widely exist in member-level predictive models in industry, e.g., feed ranking model, ads targeting model, and subscription propensity model. Spending tens of hours in model interpretation becomes a significant bottleneck in these modeling pipelines: On one hand, it is likely to cause huge delays in post-hoc model diagnosis via important feature analysis, increasing the risks of incorrect model implementations and untimely model iterations. On the other hand, it can also lead to long waiting time in preparing actionable items for model end users (e.g., marketing team in subscription propensity model) based on feature reasoning \cite{yang2021intellige}, and as a result, the end users may not take appropriate actions in a timely manner, which can negatively impact the company's revenue.

In this paper, we conduct a thorough inspection into the TreeSHAP algorithm, with the focus on improving its computational efficiency for large size of samples to be explained. In this, we take the number of samples into consideration in our time complexity analysis of TreeSHAP algorithm. We propose two new algorithms - Fast TreeSHAP v1 and Fast TreeSHAP v2. Fast TreeSHAP v1 is built based on TreeSHAP by redesigning the ways of calculating certain computationally-expensive components. In practice, Fast TreeSHAP v1 is able to consistently boost the computational speed by $\sim$1.5x with the same memory cost as in TreeSHAP. Fast TreeSHAP v2 is built based on Fast TreeSHAP v1 by further pre-computing certain computationally-expensive components which are independent with the samples to be explained. It is able to largely reduce the time complexity when calculating SHAP values, leading to 2-3x faster computational speed than TreeSHAP in practice with just a small sacrifice on memory. The pre-computation step in Fast TreeSHAP v2 enables its suitability in multi-time model interpretations, where the model is pre-trained and remains unchanged, and new scoring data are coming on a regular basis.

\section{Related Work}

Since the introductions of SHAP \cite{lundberg2017unified} and TreeSHAP \cite{lundberg2018consistent,lundberg2019explainable}, many related works have been done in this area. Most of them focus on the application side in a variety of areas, including medical science \cite{lundberg2018explainable,zhang2020clinically,agius2020machine}, social science \cite{ayush2020generating}, finance \cite{ariza2020explainability} and sports \cite{rommers2020machine}. There also exist a lot of exciting papers focusing on the designs of SHAP/TreeSHAP implementations \cite{wang2019designing,murdoch2019definitions,kaur2020interpreting,yang2021intellige}, as well as the theoretical justification of SHAP values \cite{janzing2020feature,kumar2020problems,sundararajan2020many,aas2019explaining,frye2020shapley,merrick2020explanation,chen2020true}.

Besides these works, only a few of them focus on the computational efficiency of SHAP/TreeSHAP: The authors of \cite{greenwell2019fastshap} developed a Python package \texttt{fastshap} to approximate SHAP values for arbitrary models by calling scoring function as few times as possible. In \cite{gupta2020shparkley}, a model-agnostic version of SHAP value approximation was implemented on Spark. Both works were not specifically designed for tree-based models, and thus the advanced polynomial time complexity may not be leveraged. The authors of \cite{maksymiuk2020shapper,komisarczyk2020treeshap} built R packages \texttt{shapper} and \texttt{treeshap} as R wrappers of SHAP Python library, which achieved comparable speed. However, no algorithmic improvements have been made. In \cite{messalas2020evaluating,messalas2019model}, MASHAP was proposed to compute SHAP values for arbitrary models in an efficient way, where an arbitrary model is first approximated by a surrogate XGBoost model, and TreeSHAP is then applied on this surrogate model to calculate SHAP values. The most related work of improving computational efficiency in TreeSHAP as far as we know is \cite{mitchell2020gputreeshap}, where the authors presented GPUTreeShap as a GPU implementation of TreeSHAP algorithm. Our work is different from GPUTreeShap as our work focuses on improving the computational complexity of the algorithm itself, while the parallel mechanism of GPU rather than the improvement of algorithm complexity has led to the speedup of GPUTreeShap. In our future work, it is possible to further combine Fast TreeSHAP v1 and v2 with GPUTreeShap.

\section{Background}
In this section, we review the definition of SHAP values \cite{lundberg2017unified}, and then walk through the derivation of TreeSHAP algorithm \cite{lundberg2018consistent,lundberg2019explainable} which directly leads to the development of Fast TreeSHAP v1 and v2.

\subsection{SHAP Values}
Let $f$ be the predictive model to be explained, and $N$ be the set of all input features. $f$ maps an input feature vector $x\in\mathbb{R}^{|N|}$ to an output $f(x)$ where $f(x)\in\mathbb{R}$ in regression and $f(x)\in[0, 1]$ in (binary) classification. SHAP values are defined as the coefficients of the additive surrogate explanation model: $g(z')=\phi_0+\sum_{i\in N}\phi_iz_i'$. Here $z'\in\{0,1\}^{|N|}$ represent a feature being observed ($z_i'=1$) or unknown ($z_i'=0$), and $\phi_i\in\mathbb{R}$ are the feature attribution values. Note that the surrogate explanation model $g(z')$ is a local approximation to the model prediction $f(x)$ given an input feature vector $x$.

As described in \cite{lundberg2017unified}, SHAP values are the single unique solution of $\phi_i$’s in the class of additive surrogate explanation models $g(z')$ that satisfies three desirable properties: local accuracy, missingness, and consistency. To compute SHAP values we denote $f_S(x)$ as the model output restricted to the feature subset $S\subset N$, and the SHAP values are then computed based on the classic Shapley values:
\begin{equation}\label{eq:shap_value}
	\phi_i=\sum_{S\subset N\backslash\{i\}}\frac{|S|!(|N|-|S|-1)!}{|N|!}(f_{S\cup\{i\}}(x)-f_S(x)).
\end{equation}
It is still remaining to define $f_S(x)$. In recent literature, there exist two main options for defining $f_S(x)$: the conditional expectation $f_S(x)=E[f(x)|x_S]=E_{x_{\bar{S}}|x_S}[f(x)]$, and the marginal expectation $f_S(x)=E_{x_{\bar{S}}}[f(x)]$, where $x_S$ is the sub-vector of $x$ restricted to the feature subset $S$, and $\bar{S}=N\backslash S$. While it is still debating which option is more appropriate \cite{janzing2020feature,kumar2020problems,sundararajan2020many,aas2019explaining,frye2020shapley,merrick2020explanation,chen2020true}, both options need exponential time in computation as Equation \ref{eq:shap_value} considers all possible subsets in $N$.

\subsection{SHAP Values for Trees}
For a tree-based model $f$, we note that it is sufficient to investigate the ways to calculate SHAP values on a single tree, since the SHAP values of tree ensembles equal the sums of SHAP values of its individual trees according to the additivity property of SHAP values \cite{lundberg2018consistent,lundberg2019explainable}. The authors of \cite{lundberg2018consistent,lundberg2019explainable} define a conditional expectation $f_S(x)=E[f(x)|x_S]$ for tree-based models. The basic idea is to calculate $f_S(x)$ by recursively following the decision path for $x$ if the split feature in the decision path is in $S$, and taking the weighted average of both branches if the split feature is not in $S$. We use the proportions of training data that flow down the left and right branches as the weights.

Algorithm \ref{alg:treeshap} proposed in \cite{lundberg2018consistent,lundberg2019explainable} (Appendix \ref{subsec:estimate_treeshap_detail}) provides the details to calculate $f_S(x)$ for tree-based models. A tree is specified as a tuple of six vectors $\{v, a, b, t, r, d\}$: $v$ contains the values of each leaf node. $a$ and $b$ represent the left and right node indexes for each internal node. $t$ contains the thresholds for each internal node, $d$ contains the feature indexes used for splitting in internal nodes, and $r$ represents the cover of each node (i.e., how many training samples fall in that node). The time complexity of calculating $f_S(x)$ in Algorithm \ref{alg:treeshap} is $O(L)$, where $L$ is the number of leaves in a tree, since we need to loop over each node in the tree. This leads to an exponential complexity of $O(MTL2^{|N|})$ for computing SHAP values for $M$ samples with a total of $T$ trees. We next show how TreeSHAP can help reduce this time complexity from exponential to polynomial.

\subsection{TreeSHAP}
TreeSHAP proposed in \cite{lundberg2018consistent,lundberg2019explainable} runs in $O(MTLD^2)$ time and $O(D^2+|N|)$ memory, where $M$ is the total number of samples to be explained, $|N|$ is the number of features, $T$ is the number of trees, $D$ is the maximum depth of any tree, and $L$ is the maximum number of leaves in any tree. The intuition of the polynomial time algorithm is to recursively keep track of the proportion of all possible feature subsets that flow down into each leaf node of the tree, which is similar to running Algorithm \ref{alg:treeshap} simultaneously for all $2^{|N|}$ feature subsets. We recommend readers to check \cite{lundberg2018consistent,lundberg2019explainable} for the algorithm details. In this paper, we present our understanding of the derivation of the TreeSHAP algorithm, which also leads to our proposed Fast TreeSHAP algorithms with computational advantage.

We introduce some additional notations. Assume in a given tree $\{v, a, b, t, r, d\}$, there are $K$ leaves with values $v_1$ to $v_K$, corresponding to $K$ paths $P_1$ to $P_K$, where path $P_k$ is the set of internal nodes starting from the root node and ending at the leaf node $v_k$ (leaf node $v_k$ is not included in the path). We use $j\in P_k$ to denote the $j$th internal node in path $P_k$ hereafter. We also use $D_k=\{d_j:j\in P_k\}$ to denote the feature set used for splitting in internal nodes in path $P_k$. Moreover, we denote the feature subspaces restricted by the thresholds along the path $P_k$ as $\{T_{kj}:j\in P_k\}$, where $T_{kj}=(-\infty, t_{kj}]$ if node $j$ connects to its left child in path $P_k$ or $T_{kj}=(t_{kj}, \infty)$ if node $j$ connects to its right child in path $P_k$ ($\{t_{kj}:j\in P_k\}$ are the thresholds for internal nodes in path $P_k$), and we also denote the covering ratios along the path $P_k$ as $\{R_{kj}:j\in P_k\}$, where $R_{kj}=r_{k, j+1}/r_{kj}$ ($\{r_{kj}:j\in P_k\}$ are the covers of internal nodes in path $P_k$). The formula of $f_S(x)$ in Algorithm \ref{alg:treeshap} can be simplified as:
\begin{equation*}
	f_S(x)=\sum_{k=1}^K\prod_{j\in P_k, d_j\in S}\mathbbm{1}\{x_{d_j}\in T_{kj}\}\cdot\prod_{j\in P_k, d_j\notin S}R_{kj}\cdot v_k,
\end{equation*}
where $\mathbbm{1}\{\cdot\}$ is an indicator function. Let $W_{k,S}:=\prod_{j\in P_k, d_j\in S}\mathbbm{1}\{x_{d_j}\in T_{kj}\}\cdot\prod_{j\in P_k, d_j\notin S}R_{kj}$, i.e., $W_{k, S}$ is the "proportion" of subset $S$ that flows down into the leaf node $v_k$, then $f_S(x)=\sum_{k=1}^K W_{k,S}v_k$. Plugging it into Equation \ref{eq:shap_value} leads to the SHAP value $\phi_i=\sum_{S\subset N\backslash\{i\}}\frac{|S|!(|N|-|S|-1)!}{|N|!}\sum_{k=1}^K(W_{k, S\cup\{i\}}-W_{k, S})v_k$.
\begin{theorem}\label{theorem:1}
	$\phi_i$ can be computed by only considering the paths which contain feature $i$ and all the subsets within these paths (i.e., instead of considering all $2^{|N|}$ subsets). Specifically,
	\begin{equation}\label{eq:treeshap_1}
		\begin{split}
			\phi_i=&\sum_{k:i\in D_k}\left[\sum_{m=0}^{|D_k|-1}\frac{m!(|D_k|-m-1)!}{|D_k|!}\sum_{S\subset D_k\backslash\{i\}, |S|=m}\left(\prod_{j\in P_k, d_j\in S}\mathbbm{1}\{x_{d_j}\in T_{kj}\}\prod_{j\in P_k, d_j\in D_k\backslash(S\cup\{i\})}R_{kj}\right)\right]\\
			&\cdot\left(\prod_{j\in P_k, d_j=i}\mathbbm{1}\{x_{d_j}\in T_{kj}\}-\prod_{j\in P_k, d_j=i}R_{kj}\right)v_k,
		\end{split}
	\end{equation}
\end{theorem}
The proof of Theorem \ref{theorem:1} can be found in Appendix \ref{subsec:proof_theorem_1}. For convenience, we call $\frac{m!(|D_k|-m)!}{(|D_k|+1)!}$ the "Shapley weight" for subset size $m$ and path $P_k$. We point out that the TreeSHAP algorithm is exactly built based on Theorem \ref{theorem:1}. Specifically, the EXTEND method in TreeSHAP alogithm keeps track of the sum of the "proportion"s of all subsets that flow down into a certain leaf node weighted by its Shapley weight for each possible subset size. When descending a path, EXTEND is called repeatedly to take a new feature in the path and add its contribution to the sum of the "proportion"s of all feature subsets of size $1, 2, \cdots$ up to the current depth. At the leaf node, EXTEND reaches the sequence of values $\frac{m!(|D_k|-m)!}{(|D_k|+1)!}\sum_{S\subset D_k, |S|=m}(\prod_{j\in P_k, d_j\in S}\mathbbm{1}\{x_{d_j}\in T_{kj}\}\prod_{j\in P_k, d_j\in D_k\backslash S}R_{kj})$ for each possible subset size $m=1,\cdots,|D_k|$. The UNWIND method in TreeSHAP algorithm is used to undo a previous call to EXTEND, i.e., to remove the contribution of a feature previously added via EXTEND, and is indeed commutative with EXTEND. UNWIND can be used when duplicated features are encountered in the path, or at the leaf node when calculating the contribution of each feature in the path to SHAP values according to Equation \ref{eq:treeshap_1}.

\section{Fast TreeSHAP}\label{sec:fast_treeshap}
We now further simplify Equation \ref{eq:treeshap_1} for an even faster computational speed. Consider a subset $S_k\subset D_k$ which consists of all the features in $D_k$ not satisfying the thresholds along the path $P_k$, i.e., $S_k=\{d_j: j\in P_k,\,x_{d_j}\notin T_{kj}\}$. Also define $U_{m,D_k,C}:=\frac{m!(|D_k|-m)!}{(|D_k|+1)!}\sum_{S\subset C, |S|=m}\prod_{j\in P_k, d_j\in C\backslash S}R_{kj}$ for $m=0,\cdots,|C|$, where $C\subset D_k$ is a subset of $D_k$. We see that when $C=D_k\backslash S_k$, $U_{m, D_k, D_k\backslash S_k}$ can be interpreted as the sum of the "proportion"s of all subsets that flow down into a leaf node $v_k$ (each feature in the subsets must satisfy the thresholds along the path $P_k$) weighted by its Shapley weight for subset size $m$. Finally, we define $U_{D_k, C}:=\sum_{m=0}^{|C|}\frac{|D_k|+1}{|D_k|-m}U_{m,D_k,C}$. Theorem \ref{theorem:2} further simplifies Equation \ref{eq:treeshap_1} (proof in Appendix \ref{subsec:proof_theorem_2}):
\begin{theorem}\label{theorem:2}
	$\phi_i$ can be computed by only considering the subsets within the paths where each feature in the subsets satisfies the thresholds along the path. Specifically,
	\begin{equation}\label{eq:treeshap_2}
		\begin{split}
		\phi_i=&\sum_{k:i\in D_k, i\notin S_k}\left[U_{D_k, D_k\backslash (\{i\}\cup S_k)}\prod_{j\in P_k, d_j\in S_k}R_{kj}(1-\prod_{j\in P_k, d_j=i}R_{kj})\right]v_k\\
		&-\sum_{k:i\in D_k, i\in S_k}\left[U_{D_k, D_k\backslash S_k}\prod_{j\in P_k, d_j\in S_k}R_{kj}\right]v_k.
		\end{split}
	\end{equation}
\end{theorem}

\subsection{Fast TreeSHAP v1 Algorithm}\label{subsec:fast_treeshap_v1}
We propose Fast TreeSHAP v1 algorithm based on Theorem \ref{theorem:2} which runs in $O(MTLD^2)$ time and $O(D^2+|N|)$ memory. This computational complexity looks the same as in the original TreeSHAP. However, we will show in section \ref{subsubsec:fast_treeshap_v1_complexity} and \ref{sec:evaluation} that the average running time can be largely reduced.

In Fast TreeSHAP v1 algorithm (Algorithm \ref{alg:fast_treeshap_v1} in Appendix \ref{subsec:fast_treeshap_v1_detail}), we follow the similar algorithm setting as in the original TreeSHAP algorithm, where both EXTEND and UNWIND methods are being used. The EXTEND method is used to keep track of $U_{m, D_k, D_k\backslash S_k}$ for $m=0, \cdots, |D_k|-|S_k|$. Remind that $U_{m, D_k, D_k\backslash S_k}$ is the sum of the "proportion"s of all subsets that flow down into a leaf node $v_k$ \textbf{(each feature in the subsets must satisfy the thresholds along the path $P_k$)} weighted by its Shapley weight for subset size $m$. Compared with the EXTEND method in the original TreeSHAP algorithm, the main difference is the constraint applied on these subsets (highlighted in bold), which largely reduces the number of subset sizes to be considered. Specifically, in Fast TreeSHAP v1, when descending a path, EXTEND is called only when a new feature in the path satisfies the threshold, and then its contribution to the sum of "proportion"s of all feature subsets of size $1,2,\cdots$ up to the number of features satisfying the thresholds until the current depth is added. When reaching the leaf node, the number of possible subset sizes considered by EXTEND is $|D_k|-|S_k|$ in Fast TreeSHAP v1 rather than $|D_k|$ in the original TreeSHAP. The UNWIND method is still used to undo a previous call to EXTEND. Specifically, it is used when duplicated features are encountered in the path or when calculating $U_{m, D_k, D_k\backslash(\{i\}\cup S_k)}$ for $i\in D_k\backslash S_k$, $m=0, \cdots, |D_k|-|S_k|-1$ at the leaf node. Besides EXTEND and UNWIND, we also keep track of the product of covering ratios of all features not satisfying the thresholds along the path, i.e., $\prod_{j\in P_k, d_j\in S_k}R_{kj}$ in Equation \ref{eq:treeshap_2}, which is trivial.

\subsubsection{Complexity Analysis}\label{subsubsec:fast_treeshap_v1_complexity}
In the original TreeSHAP, the complexity of EXTEND and UNWIND is bounded by $O(D)$, since both of them need to loop over the number of possible subset sizes, which equals $|D_k|$ in path $P_k$. At each internal node, EXTEND is called once, while at each leaf node, UNWIND is called $D$ times to update SHAP values for each of the $D$ features in the path. This leads to a complexity of $O(LD^2)$ for the entire tree because the work done at the leaves dominates the work at the internal nodes.

In Fast TreeSHAP v1, both EXTEND and UNWIND need to loop over the number of possible subset sizes under the constraint on subset (highlighted in bold), which is $|D_k|-|S_k|$ in path $P_k$. Thus, although the complexity of EXTEND and UNWIND is still bounded by $O(D)$, the average running time can be reduced to $50\%$, which equals the average ratio between the number of possible subset sizes under constraint and the number of all possible subset sizes, i.e., $E[(|D_k|-|S_k|)/|D_k|]\approx \sum_{i=0}^DC_D^i(D-i)/(2^DD)\approx50\%$. Moreover, according to Equation \ref{eq:treeshap_2}, at the leaf node of Path $P_k$, UNWIND is called $|D_k|-|S_k|$ times for each of the $|D_k|-|S_k|$ features satisfying the thresholds in the path, and only once for all other features in the path. Therefore, although the number of times UNWIND being called is still bounded by $O(D)$, the actual number can also be lowered by $50\%$ on average. As a result, although we still have the complexity of $O(LD^2)$ for the entire tree, the average running time can be reduced to $50\%\times50\%=25\%$ compared with the original TreeSHAP. Finally, the complexity is $O(MTLD^2)$ for the entire ensemble of $T$ trees and $M$ samples to be explained, with the running time reduced to $25\%$ on average compared with the original TreeSHAP.

\subsection{Fast TreeSHAP v2 Algorithm}\label{subsec:fast_treeshap_v2}
We propose Fast TreeSHAP v2 algorithm that runs in $O(TL2^DD+MTLD)$ time and $O(L2^D)$ memory. For balanced trees it becomes $O(TL^2D+MTLD)$ time and $O(L^2)$ memory. Compared with $O(MTLD^2)$ time and $O(D^2+|N|)$ memory in original TreeSHAP and Fast TreeSHAP v1, Fast TreeSHAP v2 outperforms in computational speed when the number of samples $M$ exceeds $2^{D+1}/D$, where $D$ is the maximum depth of any tree (more details in the next paragraph). Fast TreeSHAP v2 has a stricter restriction on tree size due to memory concerns. In practice, it works well for trees with maximum depth as large as 16 in an ordinary laptop, which covers most of the use cases of tree-based models. We discuss the memory concerns in detail in Sections \ref{subsec:fast_treeshap_summary} and \ref{sec:evaluation}.

The design of Fast TreeSHAP v2 algorithm is inspired by Fast TreeSHAP v1 algorithm. Recall that the loops of UNWIND at the leaves dominate the complexity of Fast TreeSHAP v1 algorithm, where the length of each loop is $O(D)$, and each call to UNWIND also takes $O(D)$ time, resulting in $O(D^2)$ time complexity at each leaf node. While looping over each of the $D$ features in the path is inevitable in updating SHAP values at the leaf node (i.e., the loop of length $O(D)$ is necessary), our question is: Is it possible to get rid of calling UNWIND? From Equation \ref{eq:treeshap_2} we see that, the ultimate goal of calling UNWIND is to calculate $U_{D_k, D_k\backslash S_k}$ and $U_{D_k, D_k\backslash(\{i\}\cup S_k)}$ for $i\in D_k\backslash S_k$. We also note that for different samples to be explained, although $S_k$ may vary from sample to sample, all the possible values of $U_{D_k, D_k\backslash S_k}$ and $U_{D_k, D_k\backslash(\{i\}\cup S_k)}$ for $i\in D_k\backslash S_k$ fall in the set $\{U_{D_k, C}:C\subset D_k\}$ with size $2^{D_k}$. Therefore, a natural idea to reduce the computational complexity is, instead of calling UNWIND to calculate $U_{D_k, D_k\backslash S_k}$ and $U_{D_k, D_k\backslash(\{i\}\cup S_k)}$ for $i\in D_k\backslash S_k$ every time we explain a sample, we can pre-compute all the values in the set $\{U_{D_k, C}:C\subset D_k\}$ which only depend on the tree itself, and then extract the corresponding value when looping over features at leaf nodes to calculate $\phi_i$ for each specific sample to be explained. In fact, what we just proposed is to trade space complexity for time complexity. This should significantly save computational efforts when there exist redundant computations of $U_{D_k, C}$ across samples, which generally happens when $M>2^{D+1}/D$ (For each sample, around $D/2$ $U_{D_k, C}$’s should be calculated for each path, thus on average $MD/2$ calculations should be taken for $M$ samples). This commonly occurs in a moderate-sized dataset, e.g., $M>22$ when $D=6$, $M>205$ when $D=10$, and $M>2341$ when $D=14$. We show the appropriateness of trading space complexity for time complexity in practice in Section \ref{sec:evaluation}.

We split Fast TreeSHAP v2 algorithm into two parts: \textbf{Fast TreeSHAP Prep} and \textbf{Fast TreeSHAP Score}. Fast TreeSHAP Prep (Algorithm \ref{alg:fast_treeshap_v2_prep} in Appendix \ref{subsec:fast_treeshap_v2_detail}) calculates the sets $\{U_{D_k, C}:C\subset D_k\}$ for all $D_k$’s in the tree, and Fast TreeSHAP Score (Algorithm \ref{alg:fast_treeshap_v2_score} in Appendix \ref{subsec:fast_treeshap_v2_detail}) calculates $\phi_i$’s for all samples to be explained based on the pre-computed $\{U_{D_k, C}:C\subset D_k\}$.  The main output of Fast TreeSHAP Prep is $S$, an $L\times2^D$ matrix where each row records the values in $\{U_{D_k, C}:C\subset D_k\}$ for one path $P_k$. To calculate $S$, similar to the original TreeSHAP and Fast TreeSHAP v1, both EXTEND and UNWIND methods are used. The EXTEND method keeps track of $\{U_{m, D_k, C}: m=0, \cdots, |C|\}$ for all possible subsets $C\subset D_k$ simultaneously, and the UNWIND method undoes a previous call to EXTEND when duplicated features are encountered in the path. At the leaf node, $U_{D_k, C}$ is obtained by summing up $\{U_{m, D_k, C}\}$ across $m$ for all possible subsets $C\subset D_k$ simultaneously. In Fast TreeSHAP Score, given a feature vector $x$, we need to find out its corresponding $S_k$, i.e., the feature subset within path $P_k$ where each feature satisfies the thresholds along the path, and then extract the corresponding value of $U_{D_k, D_k\backslash S_k}$ and $U_{D_k, D_k\backslash(\{i\}\cup S_k)}$ for $i\in D_k\backslash S_k$ from pre-computed $S$.

\subsubsection{Complexity Analysis}
In Fast TreeSHAP Prep, the time complexities of both EXTEND at the internal node and $S$ calculation at the leaf node are bounded by $O(2^DD)$, where $O(2^D)$ comes from the number of possible subsets within each path, and $O(D)$ comes from the number of possible subset sizes. Thus the time complexity is $O(TL2^DD)$ for the entire ensemble of $T$ trees. Note that this time complexity is independent with the number of samples to be explained, thus this entire part can be pre-computed, and matrix $S$ can be stored together with other tree properties to facilitate future SHAP value calculation. The space complexity is dominated by $S$, which is $O(L2^D)$. Note that this complexity is for one tree. In practice, there are two ways to achieve this complexity for ensemble of $T$ trees: i). Sequentially calculate $S$ for each tree, and update SHAP values for all samples immediately after one $S$ is calculated. ii). Pre-calculate $S$ for all trees and store them in the local disk, and sequentially read each $S$ into memory and update SHAP values for all samples accordingly.

In Fast TreeSHAP Score, it takes $O(1)$ time at each internal node to figure out $S_k$, and $O(D)$ time at each leaf node to loop over each of the $D$ features in the path to extract its corresponding value from $S$ (It takes $O(1)$ time to look up in $S$). Therefore, the loops at the leaves dominate the complexity of Fast TreeSHAP Score, which is $O(D)$. Finally, the complexity is $O(MTLD)$ for the entire ensemble of $T$ trees and $M$ samples to be explained. Compared with $O(MTLD^2)$ complexity in the original TreeSHAP and Fast TreeSHAP v1, this is a $D$-time improvement in computational complexity.

\subsection{Fast TreeSHAP Summary}\label{subsec:fast_treeshap_summary}
Table \ref{table:treeshap_summary} summarizes the time and space complexities of each variant of TreeSHAP algorithm ($M$ is the number of samples to be explained, $|N|$ is the number of features, $T$ is the number of trees, $L$ is the maximum number of leaves in any tree, and $D$ is the maximum depth of any tree).

\begin{table}[h]
	\caption{Summary of computational complexities of TreeSHAP algorithms.}\label{table:treeshap_summary}
	\centering
	\begin{tabular}{lll}
		\toprule
		\textbf{TreeSHAP Version} & \textbf{Time Complexity} & \textbf{Space Complexity} \\
		\hline
		Original TreeSHAP & $O(MTLD^2)$ & $O(D^2+|N|)$ \\
		\hline
		Fast TreeSHAP v1 & $O(MTLD^2)$\footnote{Average running time is reduced to $25\%$ of original TreeSHAP.} & $O(D^2+|N|)$ \\
		\hline
		Fast TreeSHAP v2 (general case) & $O(TL2^DD+MTLD)$ & $O(L2^D)$ \\
		Fast TreeSHAP v2 (balanced trees) & $O(TL^2D+MTLD)$ & $O(L^2)$ \\
		\bottomrule
		\multicolumn{3}{l}{\footnotesize{$^1$Average running time is reduced to $25\%$ of original TreeSHAP.}}
	\end{tabular}
\end{table}

Fast TreeSHAP v1 strictly outperforms original TreeSHAP in average running time and performs comparably with original TreeSHAP in space allocation. Thus we recommend to at least replace original TreeSHAP with Fast TreeSHAP v1 in any tree-based model interpretation use cases.

We consider two scenarios in model interpretation use cases to compare Fast TreeSHAP v1 and v2:
\begin{itemize}
	\item
	One-time usage: We explain all the samples for once, which usually occurs in ad-hoc model diagnosis. In this case, as mentioned in Section \ref{subsec:fast_treeshap_v2}, Fast TreeSHAP v2 is preferred when $M>2^{D+1}/D$ (commonly occurs in a moderate-sized dataset, as most tree-based models produce trees with depth $\leq16$). Also, Fast TreeSHAP v2 is under a stricter memory constraint: $O(L2^D)<memory\:tolerance$. For reference, for double type $L\times2^D$ matrix $S$ (assume in complete balanced trees, i.e., $L=2^D$), its space allocation is 32KB for $D=6$, 8MB for $D=10$, and 2GB for $D=14$. In practice, when $D$ becomes larger, it becomes harder to build a complete balanced tree, i.e., $L$ will be much smaller than $2^D$, leading to a much smaller memory allocation than the theoretical upbound. We will see this in Section \ref{sec:evaluation}.
	\item
	Multi-time usage: We have a stable model in the backend, and we receive new data to be scored on a regular basis. This happens in most of the use cases of predictive modeling in industry, where the model is trained in a monthly/yearly frequency but the scoring data are generated in a daily/weekly frequency. One advantage of Fast TreeSHAP v2 is that it is well-suited for this multi-time usage scenario. In Fast TreeSHAP v2, we only need to calculate $S$ once and store it in the local disk, and read it when new samples are coming, which leads to $D$-time computational speedup over Fast TreeSHAP v1.
\end{itemize}

\section{Evaluation}\label{sec:evaluation}
We train different sizes of random forest models for evaluation on a list of datasets in Table \ref{table:datasets}, with the goal of evaluating a wide range of tree ensembles representative of different real-world settings. While the first three datasets Adult \cite{kohavi1996scaling}, Superconductor \cite{hamidieh2018data}, and Crop \cite{khosravi2019random,khosravi2018msmd} in Table \ref{table:datasets} are publicly available, we also include one LinkedIn internal dataset “Upsell” to better illustrate the TreeSHAP implementation in industry. The Upsell dataset is used to predict how likely each LinkedIn customer is to purchase more Recruiters products by using features including product usage, recruiter activity, and company attributes. For each dataset, we fix the number of trees to be 100, and we train a small, medium, large, and extra-large model variant by setting the maximum depth of trees to be 4, 8, 12, and 16 respectively. Other hyperparameters in the random forest are left as default. Summary statistics for each model variant is listed in Table \ref{table:models} in Appendix \ref{subsec:models_table}.

\begin{table}[h]
	\caption{Datasets.}\label{table:datasets}
	\centering
	\begin{tabular}{lrrrrr}
		\toprule
		\textbf{Name} & \textbf{$\#$ Instances} & \textbf{$\#$ Attributes} & \textbf{$\#$ Attributes} & \textbf{Task} & \textbf{Classes} \\
		& & \textbf{(Original)} & \textbf{(One-Hot)} & & \\
		\midrule
		Adult\cite{kohavi1996scaling} & 48,842 & 14 & 64 & Classification & 2 \\
		Superconductor\cite{hamidieh2018data} & 21,263 & 81 & 81 & Regression & - \\
		Crop\cite{khosravi2019random,khosravi2018msmd} & 325,834 & 174 & 174 & Classification & 7 \\
		Upsell & 96,120 & 169 & 182 & Classification & 2 \\
		\bottomrule
	\end{tabular}
\end{table}

We compare the execution times of Fast TreeSHAP v1 and v2 against the existing TreeSHAP implementation in the open source SHAP package (\url{https://github.com/slundberg/shap}). For fair comparison, we directly modify the C file \texttt{treeshap.h} in SHAP package to incorporate both Fast TreeSHAP v1 and v2. All the evaluations were run on a single core in Azure Virtual Machine with size Standard\_D8\_v3 (8 cores and 32GB memory). We ran each evaluation on 10,000 samples. In Table \ref{table:execution_time}, results are averaged over 5 runs and standard deviations are also presented. To justify the correctness of Fast TreeSHAP v1 and v2, in each run we also compare the calculated SHAP values from Fast TreeSHAP v1 and v2 with SHAP values from the original TreeSHAP, and the maximal element-wise difference we observed during the entire evaluation process is $\sim10^{-13}$, which is most likely the numerical error.
\begin{table}[t]
	\caption{TreeSHAP vs Fast TreeSHAP v1 vs Fast TreeSHAP v2 - 10,000 samples (s.d. in parathesis).}\label{table:execution_time}
	\centering
	\begin{tabular}{lrrrrr}
		\toprule
		\textbf{Model} & \textbf{Original} & \textbf{Fast Tree-} & \textbf{Speedup} & \textbf{Fast Tree-} & \textbf{Speedup} \\
		& \textbf{TreeSHAP (s)} & \textbf{SHAP v1 (s)} & & \textbf{SHAP v2 (s)} & \\
		\midrule
		Adult-Small & 2.40 (0.03) & 2.11 (0.02) & \textbf{1.14} & 1.30 (0.04) & \textbf{1.85} \\
		Adult-Med & 61.04 (0.61) & 44.09 (0.61) & \textbf{1.38} & 26.62 (1.08) & \textbf{2.29} \\
		Adult-Large & 480.33 (3.60) & 333.94 (4.20) & \textbf{1.44} & 161.43 (3.95) & \textbf{2.98} \\
		Adult-xLarge & 1805.54 (13.75) & 1225.20 (8.97) & \textbf{1.47} & 827.62 (16.17) & \textbf{2.18} \\
		\midrule
		Super-Small & 2.50 (0.03) & 2.04 (0.02) & \textbf{1.23} & 1.28 (0.08) & \textbf{1.95} \\
		Super-Med & 89.93 (3.58) & 60.04 (3.56) & \textbf{1.50} & 35.65 (2.06) & \textbf{2.52} \\
		Super-Large & 1067.18 (10.52) & 663.02 (5.79) & \textbf{1.61} & 384.14 (4.78) & \textbf{2.78} \\
		Super-xLarge & 3776.44 (28.77) & 2342.44 (35.23) & \textbf{1.61} & 1988.48 (15.19) & \textbf{1.90} \\
		\midrule
		Crop-Small & 3.53 (0.07) & 2.90 (0.04) & \textbf{1.22} & 3.15 (0.02) & \textbf{1.12} \\
		Crop-Med & 69.88 (0.71) & 50.13 (0.91) & \textbf{1.39} & 34.57 (1.49) & \textbf{2.02} \\
		Crop-Large & 315.27 (6.37) & 216.05 (8.64) & \textbf{1.46} & 130.66 (3.80) & \textbf{2.41} \\
		Crop-xLarge & 552.23 (10.37) & 385.51 (8.48) & \textbf{1.43} & 290.49 (3.19) & \textbf{1.90} \\
		\midrule
		Upsell-Small & 2.80 (0.04) & 2.23 (0.05) & \textbf{1.26} & 2.20 (0.06) & \textbf{1.27} \\
		Upsell-Med & 90.64 (4.59) & 63.34 (1.82) & \textbf{1.43} & 34.02 (0.93) & \textbf{2.66} \\
		Upsell-Large & 790.83 (5.79) & 515.16 (1.66) & \textbf{1.54} & 282.98 (4.89) & \textbf{2.79} \\
		Upsell-xLarge & 2265.82 (17.44) & 1476.56 (4.20) & \textbf{1.53} & 1166.98 (15.02) & \textbf{1.94} \\
		\bottomrule
	\end{tabular}
\end{table}
We conduct pairwise comparisons between these three algorithms:
\begin{itemize}
	\item
	Original TreeSHAP vs Fast TreeSHAP v1: For medium, large, and extra-large models, we observe speedups consistently around 1.5x. We observe lower speedup (around 1.2x) for small models probably due to the insufficient computation in computationally-expensive parts. These speedups also seem much lower than the theoretical upper bound ($\sim4$) discussed in Section \ref{subsec:fast_treeshap_v1}, which is probably due to the existence of other tasks with slightly lower computational complexity in the algorithm.
	\item
	Original TreeSHAP vs Fast TreeSHAP v2: For medium and large models, we observe speedups around 2.5-3x, while the speedups drop to around 2x for extra-large models. This is because the first step Fast TreeSHAP Prep in Fast TreeSHAP v2 takes much longer time for larger models, and the execution time of Fast TreeSHAP v2 listed in Table \ref{table:execution_time} is a combination of its two steps. Later in this section, we will examine the execution times of Fast TreeSHAP Prep and Fast TreeSHAP Score separately.
	\item
	Fast TreeSHAP v1 vs Fast TreeSHAP v2: The speedups of Fast TreeSHAP v2 are consistently higher than the speedups of Fast TreeSHAP v1 except for small models, showing the effectiveness of Fast TreeSHAP v2 in improving the computational complexity. Their comparable performance for small models is also due to the insufficient computation.
\end{itemize}

\begin{table}[t]
	\caption{Fast TreeSHAP v2: Fast TreeSHAP Prep $\&$ Fast TreeSHAP Score - 10,000 samples (s.d. in parathesis).}\label{table:execution_time_2}
	\centering
	\begin{tabular}{lrrrrr}
		\toprule
		\textbf{Model} & \textbf{Original} & \textbf{Fast Tree-} & \textbf{Fast Tree-} & \textbf{Speedup} & \textbf{Space Allo-} \\
		& \textbf{TreeSHAP (s)} & \textbf{SHAP Prep (s)} & \textbf{SHAP Score (s)} & \textbf{(Large $M$)} & \textbf{cation of $S$} \\
		\midrule
		Adult-Small & 2.40 (0.03) & <0.01 (<0.01) & 1.30 (0.04) & \textbf{1.85} & 2KB \\
		Adult-Med & 61.04 (0.61) & 0.20 (0.01) & 26.42 (1.07) & \textbf{2.31} & 368KB \\
		Adult-Large & 480.33 (3.60) & 11.32 (0.14) & 150.11 (3.81) & \textbf{3.20} & 24.9MB \\
		Adult-xLarge & 1805.54 (13.75) & 268.90 (8.29) & 558.72 (7.88) & \textbf{3.23} & 955MB \\
		\midrule
		Super-Small & 2.50 (0.03) & <0.01 (<0.01) & 1.28 (0.08) & \textbf{1.95} & 2KB \\
		Super-Med & 89.93 (3.58) & 0.36 (0.01) & 35.29 (2.05) & \textbf{2.55} & 462KB \\
		Super-Large & 1067.18 (10.52) & 30.30 (0.34) & 353.84 (4.34) & \textbf{3.02} & 45.2MB \\
		Super-xLarge & 3776.44 (28.77) & 673.04 (8.35) & 1315.44 (6.84) & \textbf{2.87} & 1.76GB \\
		\midrule
		Crop-Small & 3.53 (0.07) & <0.01 (<0.01) & 3.15 (0.02) & \textbf{1.12} & 2KB \\
		Crop-Med & 69.88 (0.71) & 0.23 (0.01) & 34.34 (1.48) & \textbf{2.03} & 370KB \\
		Crop-Large & 315.27 (6.37) & 8.08 (0.09) & 122.58 (3.71) & \textbf{2.57} & 15.1MB \\
		Crop-xLarge & 552.23 (10.37) & 75.28 (2.34) & 215.21 (2.02) & \textbf{2.57}  & 323MB \\
		\midrule
		Upsell-Small & 2.80 (0.04) & <0.01 (<0.01) & 2.20 (0.06) & \textbf{1.27} & 2KB \\
		Upsell-Med & 90.64 (4.59) & 0.33 (0.01) & 33.69 (0.92) & \textbf{2.69} & 452KB \\
		Upsell-Large & 790.83 (5.79) & 24.59 (0.36) & 258.39 (4.53) & \textbf{3.06} & 33.7MB \\
		Upsell-xLarge & 2265.82 (17.44) & 442.74 (14.26) & 724.24 (7.89) & \textbf{3.13} & 996MB \\
		\bottomrule
	\end{tabular}
\end{table}

Table \ref{table:execution_time_2} shows the execution times of Fast TreeSHAP Prep and Fast TreeSHAP Score in Fast TreeSHAP v2. We see that the execution time of Fast TreeSHAP Prep is almost negligible for small models, but increases dramatically when the model size increases. This coincides with our discussions in Section \ref{subsec:fast_treeshap_summary} that for Fast TreeSHAP v2, a larger model needs a larger set of samples to offset the computational cost in Fast TreeSHAP Prep. The column “Speedup” shows the ratios between the execution times of Fast TreeSHAP Score and original TreeSHAP. For one-time usage scenarios, this column approximates the speedup when sample size $M$ is sufficiently large (i.e., Fast TreeSHAP Score dominates the execution time of Fast TreeSHAP v2). For multi-time usage scenarios, this column reflects the exact speedup when matrix $S$ is pre-computed, and newly incoming samples are being explained. We observe that the speedup increases as the model size increases, which exactly reflects the $D$-time improvement in computational complexity between Fast TreeSHAP Score and original TreeSHAP. Finally, the last column shows the space allocation of matrix $S$ which dominates the memory usage of Fast TreeSHAP v2.\footnote{Space allocation of $S$ is calculated by $\# maximum\:leaves \cdot 2^{\# maximum\:depth}\cdot8B$ for double type entries.} We can see that, although Fast TreeSHAP v2 costs more memory than the other two algorithms in theory, in practice, the memory constraint is quite loose as all the space allocations in Table \ref{table:execution_time_2} are not causing memory issues even in an ordinary laptop. Actually, the maximum depth of trees in most tree-based models in industry do not exceed 16.

\begin{figure}[t]
	\centering
	\begin{tabular}{cc}
		\includegraphics[width=0.5\textwidth]{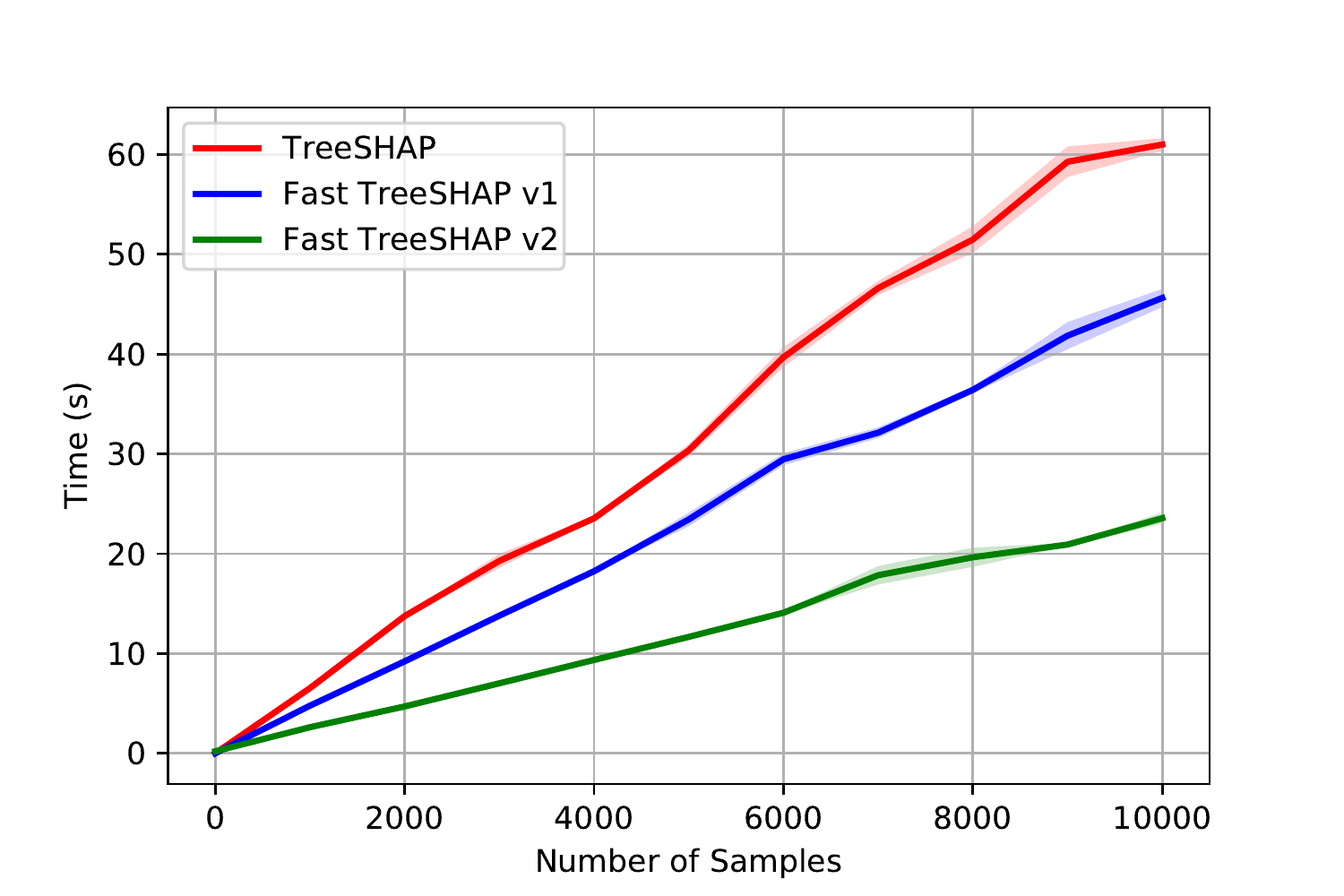} & \includegraphics[width=0.5\textwidth]{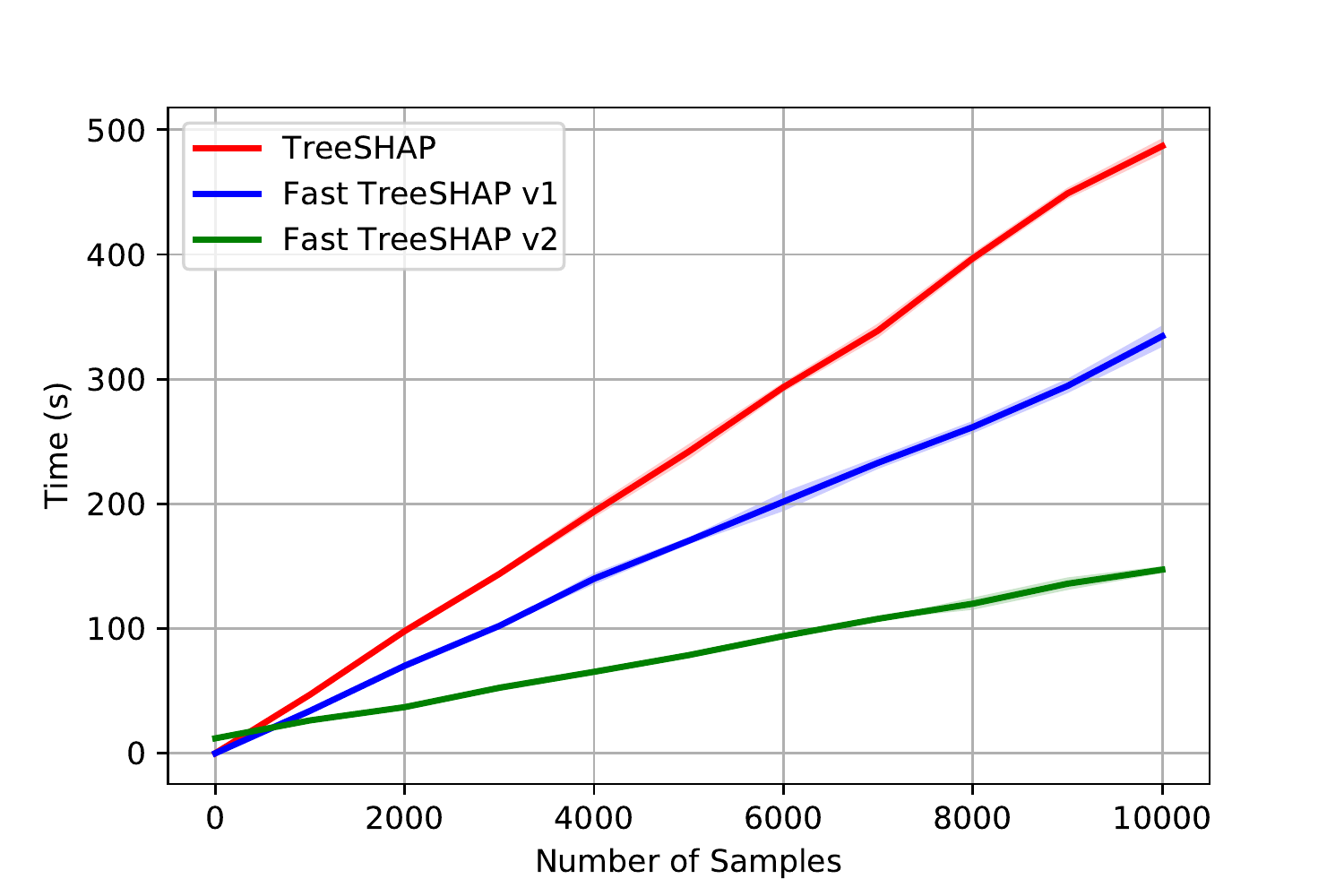}
	\end{tabular}
	\caption{Execution time v.s. number of samples. Left: Adult-Med, Right: Adult-Large.} \label{fig:execution_time}
\end{figure}

Figure \ref{fig:execution_time} plots the execution time versus number of samples for models Adult-Med and Adult-Large. $95\%$ confidence interval of the execution time is also indicated by the shaded area. Here we consider the one-time usage scenarios for a better performance showcase of the two steps in Fast TreeSHAP v2. For Adult-Med, Fast TreeSHAP v2 almost starts at the same place as the other two algorithms, since the first step Fast TreeSHAP Prep takes only 0.2s. For Adult-Large, Fast TreeSHAP Prep takes much longer time due to the larger model size, resulting in higher starting point of the green curve. However, the green curve immediately crosses the other two curves when the number of samples exceeds 500, which coincides with our previous discussions on the sample size requirement of Fast TreeSHAP v2. In both plots, Fast TreeSHAP v2 consistently performs the best while the original TreeSHAP consistently performs the worst when the number of samples exceeds a certain threshold.

We also compare the performance of original TreeSHAP, Fast TreeSHAP v1 and Fast TreeSHAP v2 on varying number of trees in Section \ref{subsec:evaluation_number_of_tree}. We find that the execution time is approximately linear in terms of the number of trees for all the three approaches, which justifies the linear term of number of trees in the theoretical complexity formulas in Table \ref{table:treeshap_summary}. We also find that the relative performance of the three approaches is consistent with the number of trees. In addition to this comparison, we also compare the performance of the three approaches on other types of tree-based models, e.g., XGBoost \cite{chen2016xgboost} model in Section \ref{subsec:evaluation_xgboost}. Overall we see that Fast TreeSHAP v1 and v2 can still achieve $\sim$1.5x and $\sim$2.5-3x speedups over the original TreeSHAP on XGBoost model.

The above evaluations are all based on 100 trees, 10,000 samples, and 1 core for fast and fair comparisons. In real life scenarios, the number of trees can be as large as several thousands, hundreds of millions of samples can be encountered in model scoring, and multi-core machines can be used to conduct parallel computing. As parallel computing is one of our future works, we just briefly discuss the potential ways to parallelize Fast TreeSHAP v1 and v2 in Appendix \ref{subsec:fast_treeshap_parallel}. Based on the proposed ways of parallel computing, we can reasonably expect that both Fast TreeSHAP v1 and v2 are able to significantly improve the computational efficiency in real life scenarios (e.g., reducing the execution time of explaining 20 million samples in a 100-core server from 15h to 5h, and reducing the execution time of explaining 320 thousand samples in a 4-core laptop from 3.5h to 1.4h). More details of the analysis of real life scenarios can be found in Appendix \ref{subsec:real_life_scenario}.

\section{Conclusion}
TreeSHAP has been widely used for explaining tree-based models due to its desirable theoretical properties and polynomial computational complexity. In this paper, we presented Fast TreeSHAP v1 and Fast TreeSHAP v2, two new algorithms to further improve the computational efficiency of TreeSHAP, with the emphasis on explaining samples with a large size. Specifically, Fast TreeSHAP v1 shrinks the computational scope for the features along each path of the tree, which is able to consistently improve the computational speed while maintaining the low memory cost. Fast TreeSHAP v2 further splits part of computationally-expensive components into pre-computation step, which significantly reduces the time complexity from $O(MTLD^2)$ to $O(TL2^DD+MTLD)$ with a small sacrifice on memory cost, and is well-suited for multi-time model explanation scenarios. We note that in practice, the computational advantage of Fast TreeSHAP v2 is achieved for sufficiently large value of $M$ (number of samples), and in the meantime $D$ (maximum depth of tree) should be upper-bounded according to the memory constraint.

As one of our future works, we are currently working on implementing the parallel computation in Fast TreeSHAP v1 and v2. Another future direction of Fast TreeSHAP is to implement it in Spark which naturally fits the environment of Hadoop clusters and the datasets stored in HDFS.

\begin{ack}
We would like to express our sincere thanks to Humberto Gonzalez, Diana Negoescu, and Wenrong Zeng for their helpful comments and feedback, and Parvez Ahammad for his support throughout this project.
\end{ack}

\newpage 
\bibliographystyle{plain}
\bibliography{Fast_TreeSHAP_reference}

\newpage
\appendix

\section{Appendix}

\subsection{Algorithm Details of Estimating $f_S(x)$}\label{subsec:estimate_treeshap_detail}
Algorithm \ref{alg:treeshap} proposed in \cite{lundberg2018consistent,lundberg2019explainable} provides the details to calculate $f_S(x)$ for tree-based models. Here a tree is specified as $\{v, a, b, t, r, d\}$, where $v$ is a vector of node values, which takes the value $internal$ for internal nodes.\footnote{A node in a tree can be either an internal node or a leaf node.} The vectors $a$ and $b$ represent the left and right node indexes for each internal node. The vector $t$ contains the thresholds for each internal node, and $d$ is a vector of indexes of the features used for splitting in internal nodes. The vector $r$ represents the cover of each node (i.e., how many data samples fall in that node from its parent).

\begin{algorithm}
	\caption{Estimating $f_S(x)$}\label{alg:treeshap}
	\begin{algorithmic}
		\Procedure{EXPVALUE}{$x$, $S$, $tree=\{v, a, b, t, r, d\}$}
		\State\Return $G(1)$ \Comment{Start recursion at the root node}
		\EndProcedure
		\Procedure{$G$}{$j$}
		\If{$v_j\neq internal$} \Comment{Check if node $j$ is a leaf}
		\State\Return $v_j$ \Comment{Return the leaf’s value}
		\Else
		\If{$d_j\in S$} \Comment{Check if the split feature is in $S$}
		\State\Return $G(a_j)$ if $x_{d_j}\leq t_j$ else $G(b_j)$ \Comment{Use value of child on the decision path}
		\Else
		\State\Return $G(a_j)r_{a_j}/r_j+G(b_j)r_{b_j}/r_j$ \Comment{Weight values of both children by coverage}
		\EndIf
		\EndIf
		\EndProcedure
	\end{algorithmic}
\end{algorithm}

\subsection{Proof of Theorem 1}\label{subsec:proof_theorem_1}
\begin{proof}
	Plugging in $f_S(x)=\sum_{k=1}^K W_{k,S}v_k$ into Equation \ref{eq:shap_value} leads to the SHAP value
	\begin{equation*}
		\phi_i=\sum_{S\subset N\backslash\{i\}}\frac{|S|!(|N|-|S|-1)!}{|N|!}\sum_{k=1}^K(W_{k, S\cup\{i\}}-W_{k, S})v_k.
	\end{equation*}
	It is easy to see that $W_{k,S\cup\{i\}}=W_{k,S}$ if $i\notin D_k$ (i.e., $\forall j\in P_k$, $d_j\neq i$). Therefore, the above equation can be simplified as
	\begin{equation*}
		\begin{split}
		\phi_i=&\sum_{S\subset N\backslash\{i\}}\frac{|S|!(|N|-|S|-1)!}{|N|!}\left[\sum_{k:i\in D_k}(W_{k, S\cup\{i\}}-W_{k, S})v_k\right]\\
		=&\sum_{k:i\in D_k}\left[\sum_{S\subset N\backslash\{i\}}\frac{|S|!(|N|-|S|-1)!}{|N|!}(W_{k, S\cup\{i\}}-W_{k, S})\right]v_k.
		\end{split}
	\end{equation*}
	Similarly, $\forall l\notin D_k$ where $l\neq i$, $\forall S\subset N\backslash\{i, l\}$, we have $W_{k,S\cup\{i, l\}}=W_{k,S\cup\{i\}}$ and $W_{k,S\cup\{l\}}=W_{k,S}$, thus $W_{k,S\cup\{i, l\}}-W_{k,S\cup\{l\}}=W_{k,S\cup\{i\}}-W_{k,S}$. Therefore, $\forall l\notin D_k$,
	\begin{equation*}
		\begin{split}
		\phi_i=&\sum_{k:i\in D_k}[\sum_{S\subset N\backslash\{i,l\}}\frac{|S|!(|N|-|S|-1)!}{|N|!}(W_{k, S\cup\{i\}}-W_{k, S})\\
		&+\sum_{S\subset N\backslash\{i,l\}}\frac{(|S|+1)!(|N|-|S|-2)!}{|N|!}(W_{k, S\cup\{i,l\}}-W_{k, S\cup\{l\}})]v_k\\
		=&\sum_{k:i\in D_k}\left[\sum_{S\subset N\backslash\{i,l\}}\left(\frac{|S|!(|N|-|S|-1)!}{|N|!}+\frac{(|S|+1)!(|N|-|S|-2)!}{|N|!}\right)(W_{k, S\cup\{i\}}-W_{k, S})\right]v_k\\
		=&\sum_{k:i\in D_k}\left[\sum_{S\subset N\backslash\{i,l\}}\frac{|S|!(|N|-|S|-2)!}{(|N|-1)!}(W_{k, S\cup\{i\}}-W_{k, S})\right]v_k.
		\end{split}
	\end{equation*}
	We repeat the above process $|N|-|D_k|$ times for each $k$, each time on a feature $l\notin D_k$, and we finally have
	\begin{equation*}
		\phi_i=\sum_{k:i\in D_k}\left[\sum_{S\subset D_k\backslash\{i\}}\frac{|S|!(|D_k|-|S|-1)!}{|D_k|!}(W_{k, S\cup\{i\}}-W_{k, S})\right]v_k.
	\end{equation*}
	Plugging in $W_{k,S}=\prod_{j\in P_k, d_j\in S}\mathbbm{1}\{x_{d_j}\in T_{kj}\}\cdot\prod_{j\in P_k, d_j\notin S}R_{kj}$ into the above equation, we have
	\begin{equation*}
		\begin{split}
			\phi_i=&\sum_{k:i\in D_k}[\sum_{S\subset D_k\backslash\{i\}}\frac{|S|!(|D_k|-|S|-1)!}{|D_k|!}(\prod_{j\in P_k, d_j\in S\cup\{i\}}\mathbbm{1}\{x_{d_j}\in T_{kj}\}\prod_{j\in P_k, d_j\in D_k\backslash(S\cup\{i\})}R_{kj}\\
			&-\prod_{j\in P_k, d_j\in S}\mathbbm{1}\{x_{d_j}\in T_{kj}\}\prod_{j\in P_k, d_j\in D_k\backslash S}R_{kj})]v_k\\
			=&\sum_{k:i\in D_k}[\sum_{S\subset D_k\backslash\{i\}}\frac{|S|!(|D_k|-|S|-1)!}{|D_k|!}(\prod_{j\in P_k, d_j\in S}\mathbbm{1}\{x_{d_j}\in T_{kj}\}\prod_{j\in P_k, d_j\in D_k\backslash(S\cup\{i\})}R_{kj})\\
			&(\prod_{j\in P_k, d_j=i}\mathbbm{1}\{x_{d_j}\in T_{kj}\}-\prod_{j\in P_k, d_j=i}R_{kj})]v_k\\
			=&\sum_{k:i\in D_k}\left[\sum_{m=0}^{|D_k|-1}\frac{m!(|D_k|-m-1)!}{|D_k|!}\sum_{S\subset D_k\backslash\{i\}, |S|=m}\left(\prod_{j\in P_k, d_j\in S}\mathbbm{1}\{x_{d_j}\in T_{kj}\}\prod_{j\in P_k, d_j\in D_k\backslash(S\cup\{i\})}R_{kj}\right)\right]\\
			&\cdot\left(\prod_{j\in P_k, d_j=i}\mathbbm{1}\{x_{d_j}\in T_{kj}\}-\prod_{j\in P_k, d_j=i}R_{kj}\right)v_k.
		\end{split}
	\end{equation*}
\end{proof}

\subsection{Proof of Theorem 2}\label{subsec:proof_theorem_2}
\begin{proof}
	Let $S_{ki}:=\{d_j: j\in P_k, x_{d_j}\notin T_{kj}, d_j\neq i\}$, i.e., $S_{ki}$ is the subset of $D_k\backslash\{i\}$ where the features in $S_{ki}$ do not satisfy the thresholds along the path $P_k$. Note that $\prod_{j\in P_k, d_j\in S}\mathbbm{1}\{x_{d_j}\in T_{kj}\}=0$ if $S\cap S_{ki}\neq\emptyset$ and $\prod_{j\in P_k, d_j\in S}\mathbbm{1}\{x_{d_j}\in T_{kj}\}=1$ otherwise, Equation \ref{eq:treeshap_1} can then be simplified as
	\begin{equation*}
		\begin{split}
			\phi_i=&\sum_{k:i\in D_k}\left[\sum_{m=0}^{|D_k|-1-|S_{ki}|}\frac{m!(|D_k|-m-1)!}{|D_k|!}\left(\sum_{S\subset D_k\backslash(\{i\}\cup S_{ki}), |S|=m}\prod_{j\in P_k, d_j\in D_k\backslash(S\cup\{i\}\cup S_{ki})}R_{kj}\right)\right]\\
			&\cdot\prod_{j\in P_k, d_j\in S_{ki}}R_{kj}\cdot\left(\prod_{j\in P_k, d_j=i}\mathbbm{1}\{x_{d_j}\in T_{kj}\}-\prod_{j\in P_k, d_j=i}R_{kj}\right)v_k.
		\end{split}
	\end{equation*}
	I.e., only subsets $S\subset D_k\backslash(\{i\}\cup S_{ki})$ are necessary to be included in the equation.
	
	Let $W_{ki}:=[\sum_{m=0}^{|D_k|-1-|S_{ki}|}\frac{m!(|D_k|-m-1)!}{|D_k|!}(\sum_{S\subset D_k\backslash(\{i\}\cup S_{ki}), |S|=m}\prod_{j\in P_k, d_j\in D_k\backslash(S\cup\{i\}\cup S_{ki})}R_{kj})]\cdot\prod_{j\in P_k, d_j\in S_{ki}}R_{kj}\cdot\left(\prod_{j\in P_k, d_j=i}\mathbbm{1}\{x_{d_j}\in T_{kj}\}-\prod_{j\in P_k, d_j=i}R_{kj}\right)$, then $\phi_i=\sum_{k:i\in D_k}W_{ki}v_k$. Remind that $S_k=\{d_j: j\in P_k,\,x_{d_j}\notin T_{kj}\}$, i.e., $S_k$ is the subset of $D_k$ where the features in $S_k$ do not satisfy the thresholds along the path $P_k$. Also remind that $U_{m,D_k,C}:=\frac{m!(|D_k|-m)!}{(|D_k|+1)!}\sum_{S\subset C, |S|=m}\prod_{j\in P_k, d_j\in C\backslash S}R_{kj}$, $m=0,\cdots,|C|$, and $U_{D_k, C}:=\sum_{m=0}^{|C|}\frac{|D_k|+1}{|D_k|-m}U_{m,D_k,C}$, where $C\subset D_k$ is a subset of $D_k$.
	
	Then it is easy to see that when $\prod_{j\in P_k, d_j=i}\mathbbm{1}\{x_{d_j}\in T_{kj}\}=0$ (i.e., $i\in S_k$), we have $S_{ki}\cup\{i\}=S_k$, and
	\begin{equation*}
		\begin{split}
			W_{ki}=&\left[\sum_{m=0}^{|D_k|-|S_k|}\frac{m!(|D_k|-m-1)!}{|D_k|!}\left(\sum_{S\subset D_k\backslash S_k, |S|=m}\prod_{j\in P_k, d_j\in D_k\backslash(S\cup S_k)}R_{kj}\right)\right]\\
			&\cdot\prod_{j\in P_k, d_j\in S_{ki}}R_{kj}\left(-\prod_{j\in P_k, d_j=i}R_{kj}\right)\\
			=&-\left[\sum_{m=0}^{|D_k|-|S_k|}\frac{|D_k|+1}{|D_k|-m}U_{m,D_k,D_k\backslash S_k}\right]\prod_{j\in P_k, d_j\in S_k}R_{kj}\\
			=&-U_{D_k, D_k\backslash S_k}\prod_{j\in P_k, d_j\in S_k}R_{kj}.
		\end{split}
	\end{equation*}
	When $\prod_{j\in P_k, d_j=i}\mathbbm{1}\{x_{d_j}\in T_{kj}\}=1$ (i.e., $i\notin S_k$), we have $S_{ki}=S_k$, and
	\begin{equation*}
		\begin{split}
			W_{ki}=&\left[\sum_{m=0}^{|D_k|-1-|S_k|}\frac{m!(|D_k|-m-1)!}{|D_k|!}\left(\sum_{S\subset D_k\backslash(\{i\}\cup S_k), |S|=m}\prod_{j\in P_k, d_j\in D_k\backslash(S\cup\{i\}\cup S_k)}R_{kj}\right)\right]\\
			&\cdot\prod_{j\in P_k, d_j\in S_k}R_{kj}\left(1-\prod_{j\in P_k, d_j=i}R_{kj}\right)\\
			=&\left[\sum_{m=0}^{|D_k|-1-|S_k|}\frac{|D_k|+1}{|D_k|-m}U_{m,D_k,D_k\backslash (\{i\}\cup S_k)}\right]\prod_{j\in P_k, d_j\in S_k}R_{kj}\left(1-\prod_{j\in P_k, d_j=i}R_{kj}\right)\\
			=&U_{D_k, D_k\backslash (\{i\}\cup S_k)}\prod_{j\in P_k, d_j\in S_k}R_{kj}(1-\prod_{j\in P_k, d_j=i}R_{kj}).
		\end{split}
	\end{equation*}	
\end{proof}

\subsection{Fast TreeSHAP v1 Algorithm Details}\label{subsec:fast_treeshap_v1_detail}
In Algorithm \ref{alg:fast_treeshap_v1}, $m$ is the path of unique features we have split on so far, and contains three attributes (We use the dot notation to access member values)\footnote{We use the same notation as in the original TreeSHAP algorithm for easy reference.}: i) $m_{j+1}.d$, the feature index for the $j$th internal node (i.e., $d_j$), ii) $m_{j+1}.z$, the covering ratio for the $j$th internal node (i.e., $R_{kj}$), iii) $m_{j+1}.o$, the threshold condition for the $j$th internal node (i.e., $\mathbbm{1}\{x_{d_j}\in T_{kj}\}$) ($m_0.d$, $m_0.z$, $m_0.o$ record the information of a dummy node). Moreover, at the $j$th internal node, $w$ is used to record $U_{m, D_{kj}, D_{kj}\backslash S_{kj}}$ ($m=0, \cdots, j-|S_{kj}|$) where $D_{kj}$ and $S_{kj}$ are subsets of $D_k$ and $S_k$ up to the $j$th internal node, and $q$ is used to record $\prod_{l\in P_{kj}, d_l\in S_{kj}}R_{kl}$ where $P_{kj}$ is the subpath of $P_k$ up to the $j$th internal node. When reaching the leaf node, $m$ will reach length $|D_k|+1$, $w$ will reach length $|D_k|-|S_k|+1$ and $q$ is always a scalar. The values $p_z$, $p_o$, and $p_i$ represent the covering ratio, the threshold condition, and the feature index of the last split. One thing to note is that the functionality of EXTEND in Algorithm \ref{alg:fast_treeshap_v1} is slightly different from what we described in Section \ref{subsec:fast_treeshap_v1}, where in addition to its functionality described in Section \ref{subsec:fast_treeshap_v1} when we encounter a feature not in $S_k$, EXTEND is also called when we encounter a feature in $S_k$ to simply update the Shapley weights.

\begin{breakablealgorithm}
	\caption{Fast TreeSHAP v1}\label{alg:fast_treeshap_v1}
	\begin{algorithmic}
		\Procedure{TreeSHAP}{$x$, $tree=\{v, a, b, t, r, d\}$}
		\State $\phi=$ array of $len(x)$ zeros
		\State RECURSE$(0, [], [], 1, 1, 1, -1)$ \Comment{Start recursion at the root node}
		\State\Return $\phi$
		\EndProcedure
		\Procedure{RECURSE}{$j, m, w, q, p_z, p_o, p_i$}
		\State $m, w, q=$ EXTEND$(m, w, q, p_z, p_o, p_i)$ \Comment{Update 
			$m, w, q$ for the feature of the last split}
		\If{$v_j\neq internal$} \Comment{Check if we are at a leaf node}
		\If{$len(m)>len(w)$}
		\State $s_0=-sum(\text{UNWIND}(m, w, q, -1)[1])$ \Comment{Pre-calculate $U_{D_k, D_k\backslash S_k}$ for $d_i$'s in $S_k$}
		\EndIf
		\For{$i\gets1$ to $len(m)-1$}
		\If{$m_i.o=0$}
		\State $\phi_{m_i.d}=\phi_{m_i.d}+s_0\cdot q\cdot v_j$ \Comment{Update contribution for feature $d_i\in S_k$}
		\Else
		\State $s=sum(\text{UNWIND}(m, w, q, i)[1])$ \Comment{Calculate $U_{D_k, D_k\backslash(\{d_i\}\cup S_k)}$ for $d_i\notin S_k$}
		\State $\phi_{m_i.d}=\phi_{m_i.d}+s\cdot q\cdot(1-m_i.z)\cdot v_j$ \Comment{Update contribution for feature $d_i\notin S_k$}
		\EndIf
		\EndFor
		\Else
		\State $h, c=(a_j, b_j)$ if $x_{d_j}\leq t_j$ else $(b_j, a_j)$ \Comment{Determine hot and cold children}
		\State $i_z=i_o=1$
		\State $k=\text{FINDFIRST}(m.d, d_j)$
		\If{$k\neq nothing$} \Comment{Undo previous update for feature $d_k$ if we see this feature again}
		\State $i_z, i_o=m_k.z, m_k.o$
		\State $m, w, q=\text{UNWIND}(m, w, q, k)$
		\EndIf
		\State RECURSE$(h, m, w, q, i_zr_h/r_j, i_o, d_j)$
		\State RECURSE$(c, m, w, q, i_zr_c/r_j, 0, d_j)$
		\EndIf
		\EndProcedure
		\Procedure{EXTEND}{$m, w, q, p_z, p_o, p_i$} \Comment{Update 
			$m, w, q$ for the feature of the last split}
		\State $l, l_w=len(m), len(w)$
		\State $m, w=copy(m), copy(w)$
		\State $m_l.(d, z, o)=(p_i, p_z, p_o)$ \Comment{Extend $m$ given the feature of the last split}
		\If{$p_o=0$}
		\State $q=q\cdot p_z$
		\For{$i\gets l_w-1$ to 0}
		\State $w_i=w_i\cdot(l-i)/(l+1)$
		\EndFor
		\Else
		\State $w_{l_w}=1$ if $l_w=0$ else 0 \Comment{Extend $w$ if the feature of the last split is not in $S_k$}
		\For{$i\gets l_w-1$ to 0}
		\State $w_{i+1}=w_{i+1}+w_i\cdot(i+1)/(l+1)$
		\State $w_i=p_z\cdot w_i\cdot(l-i)/(l+1)$
		\EndFor
		\EndIf
		\State\Return $m, w, q$
		\EndProcedure
		\Procedure{UNWIND}{$m, w, q, i$} \Comment{Undo previous update for feature $d_i$}
		\State $l, l_w=len(m)-1, len(w)-1$
		\State $m=copy(m_{0\cdots l-1})$
		\If{$i<0$}
		\State $w=copy(w)$
		\For{$j\gets l_w$ to 0}
		\State $w_j=w_j\cdot(l+1)/(l-j)$
		\EndFor
		\Else
		\If{$m_i.o=0$}
		\State $w=copy(w)$
		\For{$j\gets l_w$ to 0}
		\State $w_j=w_j\cdot(l+1)/(l-j)$
		\EndFor
		\State $q=q/m_i.z$
		\Else
		\State $n=w_{l_w}$
		\State $w=copy(w_{0\cdots l_w-1})$ \Comment{Shrink $w$ if feature $d_i\notin S_k$}
		\For{$j\gets l_w-1$ to 0}
		\State $t=w_j$
		\State $w_j=n\cdot(l+1)/(j+1)$
		\State $n=t-w_j\cdot m_i.z\cdot(l-j)/(l+1)$
		\EndFor
		\EndIf
		\For{$j\gets i$ to $l-1$}
		\State $m_j.(d, z, o)=m_{j+1}.(d, z, o)$ \Comment{Shrink $m$ given feature $d_i$}
		\EndFor
		\EndIf
		\State\Return $m, w, q$
		\EndProcedure
	\end{algorithmic}
\end{breakablealgorithm}

\subsection{Fast TreeSHAP v2 Algorithm Details}\label{subsec:fast_treeshap_v2_detail}
In Algorithm \ref{alg:fast_treeshap_v2_prep} and \ref{alg:fast_treeshap_v2_score}, $S$ is a matrix of size $L\times2^D$ where each row records the values in the set $\{U_{D_k, C}:C\subset D_k\}$ for one path $P_k$, and the subsets $C$ in each row are ordered according to the reverse ordering of binary digits.\footnote{E.g., in a 3-element set, using 0/1 to indicate absence/presence of an element, the subsets in reverse ordering of binary digits are $\{(0, 0, 0),\:(1, 0, 0),\:(0, 1, 0),\:(1, 1, 0),\:(0, 0, 1),\:(1, 0, 1),\:(0, 1, 1),\:(1, 1, 1)\}$.} $W$ is a matrix of size $2^D\times(D+1)$, where each row corresponds to a subset $C\subset D_k$ ordered according to the reverse ordering of binary digits. At the $j$th internal node along path $P_k$, each row of $W$ records $\{U_{m, D_{kj}, C}:m=0,\cdots,|C|\}$ for a subset $C\subset D_{kj}$ where $D_{kj}$ is the subset of $D_k$ up to the $j$th internal node. It is easy to see that at the leaf node of path $P_k$, summing up each row of $W$ exactly matches one row in $S$ corresponding to path $P_k$. $l_w$ is a vector of size $2^D$ which records the number of nonzero elements in each row of $W$. $e$ is a vector of the same size as $d$, which records the node indices of duplicated features along the path. $c$ is an integer which records the number of paths explored so far.

\begin{breakablealgorithm}
	\caption{Fast TreeSHAP v2: Prep}\label{alg:fast_treeshap_v2_prep}
	\begin{algorithmic}
		\Procedure{TreeSHAPPrep}{$tree=\{v,a,b,t,r,d\}$}
		\State $S=$ matrix of $L\times 2^D$ zeros
		\State $e=$ array of $len(d)$ negative ones
		\State $W=$ matrix of $2^D\times(D+1)$ zeros
		\State $l_w=$ array of $2^D$ zeros
		\State $c=0$
		\State RECURSE$(0, [], W, l_w, 1, -1)$ \Comment{Start recursion at the root node}
		\State\Return $S, e$
		\EndProcedure
		\Procedure{RECURSE}{$j,m,W,l_w,p_z,p_i$}
		\State $m,W,l_w=\text{EXTEND}(m,W,l_w,p_z,p_i)$ \Comment{Update $m, W, l_w$ for the feature of the last split}
		\If{$v_j\neq internal$} \Comment{Check if we are at a leaf node}
		\State $l=len(m)-1$
		\For{$t\gets0$ to $2^l-1$} \Comment{Calculate $\{U_{D_k, C}:C\subset D_k\}$ for the current path}
		\For{$i\gets l_w[t]$ to 0}
		\State $S[c,t]=S[c,t]+W[t,i]\cdot(l+1)/(l-i)$
		\EndFor
		\EndFor
		\State $c=c+1$
		\Else
		\State $i_z=1$
		\State $k=\text{FINDFIRST}(m.d, d_j)$
		\If{$k\neq nothing$}
		\State $e_j=k$ \Comment{Record node index of duplicated feature}
		\State $i_z=m_k.z$
		\State $m,W,l_w=\text{UNWIND}(m,W,l_w,k)$ \Comment{Undo previous update for $d_k$ if we see it again}
		\EndIf
		\State $l_w=l_w+1$
		\State RECURSE$(a_j, m, W, l_w, i_zr_{a_j}/r_j, d_j)$
		\State RECURSE$(b_j, m, W, l_w, i_zr_{b_j}/r_j, d_j)$
		\EndIf
		\EndProcedure
		\Procedure{EXTEND}{$m,W,l_w,p_z,p_i$} \Comment{Update $m, W, l_w$ for the feature of the last split}
		\State $l=len(m)$
		\State $m=copy(m)$
		\State $m_l.(d,z)=(p_i,p_z)$ \Comment{Extend $m$ given the feature of the last split}
		\If{$l=0$}
		\State $W[0,0]=1$
		\Else \Comment{Extend $W, l_w$ given the feature of the last split}
		\State $W,l_w=copy(W),copy(l_w)$
		\State $W[2^{l-1}:2^l, :]=copy(W[:2^{l-1}, :])$
		\State $l_w[2^{l-1}:2^l]=copy(l_w[:2^{l-1}])$
		\For{$t\gets0$ to $2^{l-1}-1$}
		\For{$i\gets l_w[t]-1$ to 0}
		\State $W[t,i]=W[t,i]\cdot(l-i)/(l+1)$
		\EndFor
		\EndFor
		\For{$t\gets2^{l-1}$ to $2^l-1$}
		\For{$i\gets l_w[t]-1$ to 0}
		\State $W[t,i+1]=W[t,i+1]+W[t,i]\cdot(i+1)/(l+1)$
		\State $W[t,i]=p_z\cdot W[t,i]\cdot(l-i)/(l+1)$
		\EndFor
		\EndFor
		\State $l_w[:2^{l-1}]=l_w[:2^{l-1}]-1$
		\EndIf
		\State\Return $m,W,l_w$
		\EndProcedure
		\Procedure{UNWIND}{$m,W,l_w,i$} \Comment{Undo previous update for feature $d_i$}
		\State $l=len(m)-1$
		\State $m,W,l_w=copy(m_{0\cdots l-1}),copy(W),copy(l_w)$
		\State $ind=$ empty array
		\For{$t_1\gets\{0,2^i,2\cdot2^i,3\cdot2^i,\cdots,2^l-2^i\}$}
		\For{$t_2\gets t_1$ to $t_1+2^{i-1}-1$}
		\State $ind.append(t_2)$
		\EndFor
		\EndFor
		\State $W[:2^{l-1}, :]=W[ind, :]$ \Comment{Shrink $W, l_w$ given feature $d_i$}
		\State $l_w[:2^{l-1}]=l_w[ind]$
		\For{$t\gets0$ to $2^{l-1}-1$}
		\For{$i\gets l_w[t]$ to 0}
		\State $W[t,i]=W[t,i]\cdot(l+1)/(l-i)$
		\EndFor
		\EndFor
		\For{$j\gets i$ to $l-1$}
		\State $m_j.(d,z)=m_{j+1}.(d,z)$ \Comment{Shrink $m$ given feature $d_i$}
		\EndFor
		\State\Return $m,W,l_w$
		\EndProcedure
	\end{algorithmic}
\end{breakablealgorithm}

\begin{breakablealgorithm}
	\caption{Fast TreeSHAP v2: Score}\label{alg:fast_treeshap_v2_score}
	\begin{algorithmic}
		\Procedure{TreeSHAPScore}{$x, tree=\{v,a,b,t,r,d,S,e\}$}
		\State $\phi=$ array of $len(x)$ zeros
		\State $b=$ array of $[0,1,2,2^2,\cdots,2^D]$
		\State $c=0$
		\State RECURSE$(0,[],1,1,1,-1)$ \Comment{Start recursion at the root node}
		\State\Return $\phi$
		\EndProcedure
		\Procedure{RECURSE}{$j,m,q,p_z,p_o,p_i$}
		\State $l=len(m)$
		\State $m=copy(m)$
		\State $m_l.(d,z,o)=(p_i,p_z,p_o)$ \Comment{Update $m, q$ for the feature of the last split}
		\If{$p_o=0$}
		\State $q=q\cdot p_z$
		\EndIf
		\If{$v_j\neq internal$} \Comment{Check if we are at a leaf node}
		\State $b_{sum}=0$
		\For{$i\gets1$ to $l$} \Comment{Use $b_{sum}$ to search in $\{U_{D_k, C}:C\subset D_k\}$ for the current path}
		\State $b_{sum}=b_{sum}+b_i$ if $m_i.o\neq0$
		\EndFor
		\For{$i\gets1$ to $l$}
		\If{$m_i.o=0$} \Comment{Update contribution for feature $d_i\in S_k$}
		\State $\phi_{m_i.d}=\phi_{m_i.d}-S[c,b_{sum}]\cdot q\cdot v_j$
		\Else \Comment{Update contribution for feature $d_i\notin S_k$}
		\State $\phi_{m_i.d}=\phi_{m_i.d}+S[c,b_{sum}-b_i]\cdot q\cdot(1-m_i.z)\cdot v_j$
		\EndIf
		\EndFor
		\State $c=c+1$
		\Else
		\State $i_z=i_o=1$
		\State $k=e_j$
		\If{$k\neq-1$} \Comment{Undo previous update for feature $d_k$ if we see this feature again}
		\State $i_z,i_o=m_k.z,m_k.o$
		\State $m=copy(m_{0\cdots l-1})$
		\For{$j\gets k$ to $l-1$}
		\State $m_j.(d,z,o)=m_{j+1}.(d,z,o)$
		\EndFor
		\If{$i_o=0$}
		\State $q=q/i_z$
		\EndIf
		\EndIf
		\State RECURSE$(a_j,m,q,i_zr_{a_j}/r_j,i_o\cdot\mathbbm{1}(x_{d_j}\leq t_j),d_j)$
		\State RECURSE$(b_j,m,q,i_zr_{b_j}/r_j,i_o\cdot\mathbbm{1}(x_{d_j}>t_j),d_j)$
		\EndIf
		\EndProcedure
	\end{algorithmic}
\end{breakablealgorithm}

\subsection{Additional Tables in Evaluation}\label{subsec:models_table}
Table \ref{table:models} lists the summary statistics for each model variant used in the evaluation study. We fix the number of trees to be 100, since the computational time is approximately linear in terms of the number of trees, therefore without loss of generalizability, we simply set the number of trees to be a fixed number.
\begin{table}[h]
	\caption{Models.}\label{table:models}
	\centering
	\begin{tabular}{lrrr}
		\toprule
		\textbf{Model} & \textbf{$\#$ Trees} & \textbf{Max Depth} & \textbf{$\#$ Leaves} \\
		\midrule
		Adult-Small & 100 & 4 & 1,549 \\
		Adult-Med & 100 & 8 & 14,004 \\
		Adult-Large & 100 & 12 & 57,913 \\
		Adult-xLarge & 100 & 16 & 140,653 \\
		\midrule
		Super-Small & 100 & 4 & 1,558 \\
		Super-Med & 100 & 8 & 19,691 \\
		Super-Large & 100 & 12 & 124,311 \\
		Super-xLarge & 100 & 16 & 314,763 \\
		\midrule
		Crop-Small & 100 & 4 & 1,587 \\
		Crop-Med & 100 & 8 & 13,977 \\
		Crop-Large & 100 & 12 & 37,503 \\
		Crop-xLarge & 100 & 16 & 51,195 \\
		\midrule
		Upsell-Small & 100 & 4 & 1,600 \\
		Upsell-Med & 100 & 8 & 20,247 \\
		Upsell-Large & 100 & 12 & 92,101 \\
		Upsell-xLarge & 100 & 16 & 181,670 \\
		\bottomrule
	\end{tabular}
\end{table}

\subsection{Evaluation on Varying Number of Trees}\label{subsec:evaluation_number_of_tree}
\begin{figure}[t]
	\centering
	\begin{tabular}{cc}
		\includegraphics[width=0.5\textwidth]{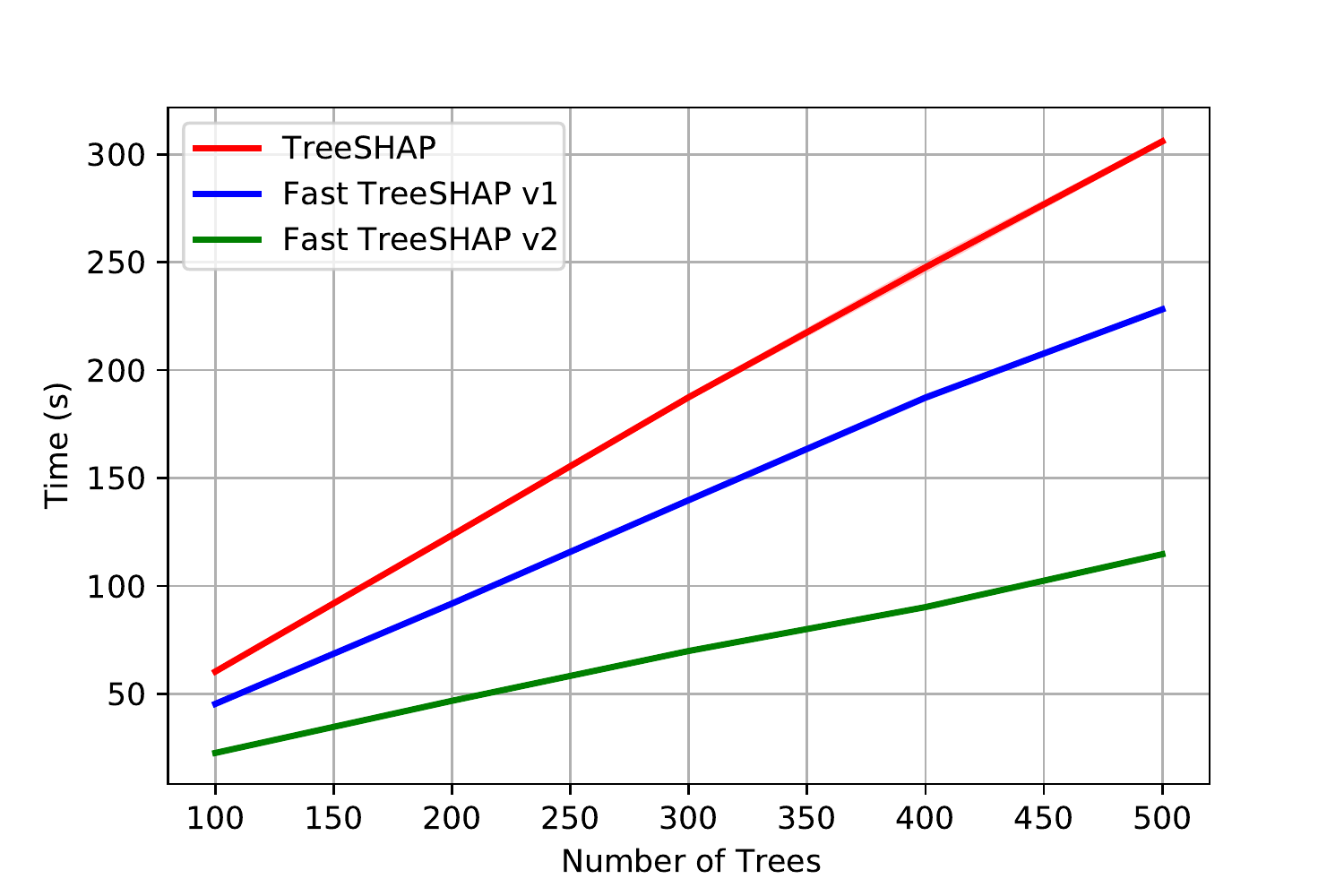} & \includegraphics[width=0.5\textwidth]{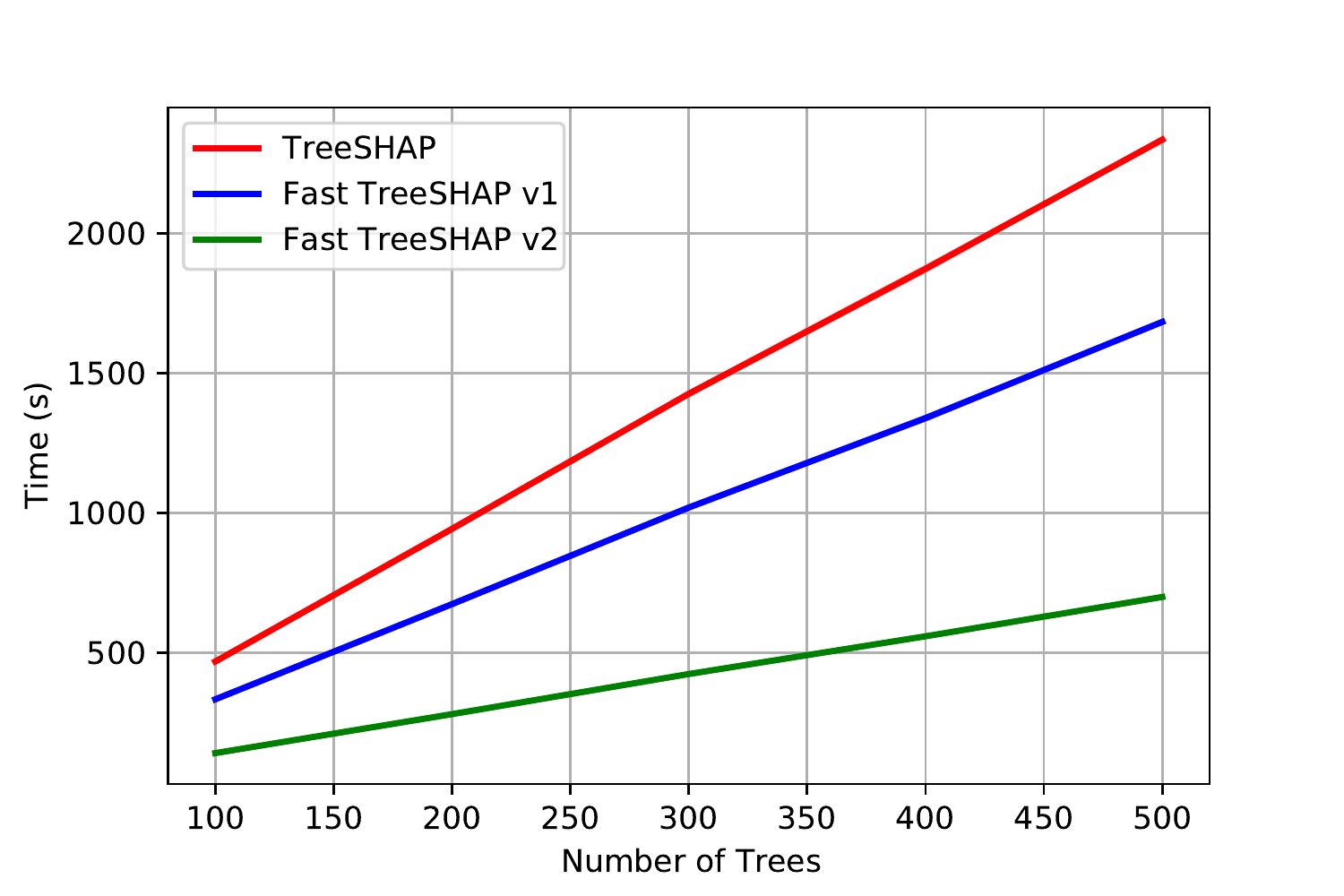}
	\end{tabular}
	\caption{Execution time v.s. number of trees. Left: Adult-Med, Right: Adult-Large.} \label{fig:execution_time_tree}
\end{figure}

We compare the performance of original TreeSHAP, Fast TreeSHAP v1 and Fast TreeSHAP v2 on varying number of trees. Figure \ref{fig:execution_time_tree} plots the execution time versus number of trees for models Adult-Med and Adult-Large. 95\% confidence interval of the execution time is also indicated by the shaded area. We find that in both plots, the execution time is approximately linear in terms of the number of trees for all the three approaches, which justifies the theoretical complexity formulas in Table \ref{table:treeshap_summary}. We also find that the relative performance of the three approaches is consistent with the number of trees, where Fast TreeSHAP v2 consistently performs the best, while original TreeSHAP consistently performs the worst.

\subsection{Evaluation on XGBoost Model}\label{subsec:evaluation_xgboost}
We compare the performance of original TreeSHAP, Fast TreeSHAP v1 and Fast TreeSHAP v2 on XGBoost \cite{chen2016xgboost} models trained on datasets Adult \cite{kohavi1996scaling} and Superconductor \cite{hamidieh2018data}. For fair comparison, all the evaluations were run on a single core in Azure Virtual Machine with size Standard\_D8\_v3 (8 cores and 32GB memory). We construct Table \ref{table:models_xgboost}, \ref{table:execution_time_xgboost}, and \ref{table:execution_time_2_xgboost} in the same format of Table \ref{table:models}, \ref{table:execution_time}, and \ref{table:execution_time_2}. Overall we see that Fast TreeSHAP v1 and v2 can still achieve $\sim$1.5x and $\sim$2.5-3x speedups than original TreeSHAP.

\begin{table}[h]
	\caption{Models.}\label{table:models_xgboost}
	\centering
	\begin{tabular}{lrrr}
		\toprule
		\textbf{Model} & \textbf{$\#$ Trees} & \textbf{Max Depth} & \textbf{$\#$ Leaves} \\
		\midrule
		Adult-Small & 100 & 4 & 1,480 \\
		Adult-Med & 100 & 8 & 7,375 \\
		Adult-Large & 100 & 12 & 21,085 \\
		Adult-xLarge & 100 & 16 & 36,287 \\
		\midrule
		Super-Small & 100 & 4 & 1,512 \\
		Super-Med & 100 & 8 & 12,802 \\
		Super-Large & 100 & 12 & 47,840 \\
		Super-xLarge & 100 & 16 & 108,010 \\
		\bottomrule
	\end{tabular}
\end{table}

\begin{table}[h]
	\caption{TreeSHAP vs Fast TreeSHAP v1 vs Fast TreeSHAP v2 - 10,000 samples (s.d. in parathesis).}\label{table:execution_time_xgboost}
	\centering
	\begin{tabular}{lrrrrr}
		\toprule
		\textbf{Model} & \textbf{Original} & \textbf{Fast Tree-} & \textbf{Speedup} & \textbf{Fast Tree-} & \textbf{Speedup} \\
		& \textbf{TreeSHAP (s)} & \textbf{SHAP v1 (s)} & & \textbf{SHAP v2 (s)} & \\
		\midrule
		Adult-Small & 2.03 (0.05) & 1.92 (0.05) & \textbf{1.06} & 0.85 (0.01) & \textbf{2.39} \\
		Adult-Med & 23.61 (0.11) & 18.83 (0.14) & \textbf{1.25} & 7.65 (0.07) & \textbf{3.09} \\
		Adult-Large & 113.77 (0.44) & 83.58 (0.55) & \textbf{1.36} & 36.98 (0.19) & \textbf{3.08} \\
		Adult-xLarge & 256.32 (1.24) & 183.07 (1.65) & \textbf{1.40} & 78.02 (0.33) & \textbf{3.29} \\
		\midrule
		Super-Small & 2.61 (0.10) & 2.42 (0.10) & \textbf{1.08} & 1.12 (0.06) & \textbf{2.33} \\
		Super-Med & 47.00 (1.64) & 33.68 (0.22) & \textbf{1.40} & 17.45 (0.12) & \textbf{2.69} \\
		Super-Large & 334.28 (12.46) & 221.38 (2.63) & \textbf{1.51} & 105.85 (2.71) & \textbf{3.16} \\
		Super-xLarge & 1124.07 (4.50) & 720.21 (7.84) & \textbf{1.56} & 445.49 (5.03) & \textbf{2.52} \\
		\bottomrule
	\end{tabular}
\end{table}

\begin{table}[h]
	\caption{Fast TreeSHAP v2: Fast TreeSHAP Prep $\&$ Fast TreeSHAP Score - 10,000 samples (s.d. in parathesis).}\label{table:execution_time_2_xgboost}
	\centering
	\begin{tabular}{lrrrrr}
		\toprule
		\textbf{Model} & \textbf{Original} & \textbf{Fast Tree-} & \textbf{Fast Tree-} & \textbf{Speedup} & \textbf{Space Allo-} \\
		& \textbf{TreeSHAP (s)} & \textbf{SHAP Prep (s)} & \textbf{SHAP Score (s)} & \textbf{(Large $M$)} & \textbf{cation of $S$} \\
		\midrule
		Adult-Small & 2.03 (0.05) & <0.01 (<0.01) & 0.85 (0.01) & \textbf{2.39} & 2KB \\
		Adult-Med & 23.61 (0.11) & 0.05 (<0.01) & 7.60 (0.07) & \textbf{3.11} & 258KB \\
		Adult-Large & 113.77 (0.44) & 1.03 (0.02) & 35.95 (0.19) & \textbf{3.16} & 14.2MB \\
		Adult-xLarge & 256.32 (1.24) & 6.76 (0.08) & 71.26 (0.33) & \textbf{3.60} & 421MB \\
		\midrule
		Super-Small & 2.61 (0.10) & <0.01 (<0.01) & 1.12 (0.06) & \textbf{2.33} & 2KB \\
		Super-Med & 47.00 (1.64) & 0.15 (0.01) & 17.30 (0.12) & \textbf{2.72} & 392KB \\
		Super-Large & 334.28 (12.46) & 7.42 (0.06) & 98.43 (2.71) & \textbf{3.40} & 27.7MB \\
		Super-xLarge & 1124.07 (4.50) & 147.00 (3.87) & 298.49 (5.03) & \textbf{3.77} & 936MB \\
		\bottomrule
	\end{tabular}
\end{table}

\subsection{Fast TreeSHAP Parallelization}\label{subsec:fast_treeshap_parallel}
Parallel computing is not the main topic of this paper and will be one of our future works. Here we just briefly discuss the potential ways to parallelize Fast TreeSHAP v1 and v2 to achieve an even higher computational speed. Fast TreeSHAP v1 can be parallelized per sample, similar to the parallelization setting in the original TreeSHAP. The parallelization setting for Fast TreeSHAP v2 is a bit complicated: For one-time usage, per-tree parallelization is recommended as long as the total memory cost across multiple threads (i.e., $\#thread\cdot O(L2^D)$) fits the entire memory, which works for most moderate-sized trees. For multi-time usage or when per-tree parallelization is not applicable, we can pre-compute $S$'s for all trees and store them in the local disk, and read them when explaining the samples. In this case, per-sample parallelization can be implemented. Specifically, we can read these $S$'s in batches with size $B$ which leads to $O(BL2^D)$ memory cost. When the corresponding tree size is small, we can increase the batch size to potentially reduce the overheads in data loading. When the tree size is large, we can decrease the batch size until it fits the memory well. Once a batch of $S$'s have been loaded into memory, we can implement per-sample parallelization to update SHAP values for all samples for this batch of trees.

\subsection{Analysis of Real Life Scenarios}\label{subsec:real_life_scenario}
The evaluation studies in Section \ref{sec:evaluation} are all based on 100 trees, 10,000 samples, and 1 core for fast and fair comparisons. In real life scenarios in industry, the number of trees is usually between 200 and 1000, hundreds of millions of samples can be encountered in model scoring, and multi-core machines can be used to conduct parallel computing. Based on the discussions in Section \ref{subsec:fast_treeshap_parallel} and the linear patterns in Figure \ref{fig:execution_time}, we can reasonably expect the linear scaling of the performance of the three TreeSHAP algorithms with respect to the number of trees, the number of samples, and the number of threads in parallel computing. Here we list two real life scenarios and estimate the improvement of Fast TreeSHAP v1 and v2 over the original TreeSHAP (The estimates may not be very accurate, and they are mostly used to provide an intuitive view of the magnitude of problems and the improvement of proposed approaches):
\begin{itemize}
	\item
	In the prediction of Premium subscriptions of LinkedIn members, 20 million LinkedIn members need to be scored every week. The model we have built is a random forest model which is refreshed every half a year. This model contains 400 trees with the maximum depth of 12. Assume we are using a server with 100 cores. We can reasonably approximate from Table \ref{table:execution_time} that the execution time of explaining the entire 20 million samples by using the original TreeSHAP is about $660\times4\times2000/100$s = 14.7h (we use the average execution time of the large models across 4 datasets) even with parallel computing on 100 cores. This computational cost is much higher than the model scoring itself! We propose using Fast TreeSHAP v2 as this is the multi-time usage scenario, where matrix $S$ can be pre-computed and stored in the server. From Table \ref{table:execution_time_2}, we can reasonably estimate that the matrix $S$ occupies about $400\times30$MB=12GB in storage (store all 400 trees together), and 30MB in memory (read each tree sequentially) when running Fast TreeSHAP v2, which should well fit the industry-level server. We can also estimate from Table \ref{table:execution_time_2} that Fast TreeSHAP v2 provides $\sim3$x speedup compared with the original TreeSHAP, which reduces the execution time of model interpretation from 14.7h to just 4.9h.
	\item
	In the crop mapping prediction \cite{khosravi2019random,khosravi2018msmd}, 320 thousand remote sensing images need to be scored on a 4-core laptop. The model we have built is similar to the model used in Section \ref{sec:evaluation}: a random forest model with 500 trees and maximum depth of 12. From Table \ref{table:execution_time}, it is reasonable to estimate that the execution time of the original TreeSHAP is about $315\times32\times5/4$s = 3.5h when parallelizing on the entire 4 cores. Fast TreeSHAP v1 can help reduce the execution time to $216\times32\times5/4$s = 2.4h. Since this is the one-time usage scenario, per-tree parallelization can be implemented for Fast TreeSHAP v2, which costs around $15\times4$MB = 60MB in memory, and can further reduce the execution time to $130\times32\times5/4$s = 1.4h.
\end{itemize}

\end{document}